\documentclass[journal]{IEEEtran}

\usepackage{ifpdf}
\usepackage{color}
\usepackage{amssymb}
\usepackage[linesnumbered,ruled,vlined]{algorithm2e}
\usepackage{cite}
\usepackage{amsmath}
\usepackage{units}
\usepackage{array}
\usepackage{stfloats}
\usepackage{url}
\usepackage{hyperref}
\usepackage{subfigure}
\usepackage{footnote}
\usepackage{multirow}
\usepackage{booktabs}
\usepackage{tabularx}
\usepackage[table,xcdraw]{xcolor}
\usepackage{threeparttable}
\usepackage{bbding}
\usepackage{makecell}

\usepackage{orcidlink}
\usepackage{pifont}

\begin{document}
\title{A Survey on End-to-End Autonomous Driving Training from the Perspectives of Data, Strategy, and Platform}

\author{Chengkai~Xu$^{\orcidlink{0009-0000-6714-3004}}$,
        Yiming~Cui$^{\orcidlink{0009-0000-3626-5711}}$,
        Jiaqi~Liu$^{\orcidlink{0000-0002-6171-6155}}$,
        Yicheng~Guo$^{\orcidlink{0009-0002-3945-2007}}$,
        Cheng~Qin$^{\orcidlink{0009-0006-0873-107X}}$, \\
        Geyuan~Zhang,
        Xinwei~Dong,
        Shiyu~Fang$^{\orcidlink{0009-0000-2884-1504}}$,
        Peng~Hang$^{\orcidlink{0000-0002-5843-0594}}$, ~\IEEEmembership{Senior Member,~IEEE,}
        and~Jian~Sun$^{\orcidlink{0000-0001-5031-4938}}$ 
\thanks{
This work was partly supported by the National Natural Science Foundation of China (52472451), the Shanghai Scientific Innovation Foundation (No.23DZ1203400), and the Fundamental Research Funds for the Central Universities.
}
\thanks{
Chengkai Xu, Yiming Cui, Yicheng Guo, Cheng Qin, Geyuan Zhang, Xinwei Dong, Shiyu Fang,  Peng Hang and Jian Sun are with the College of Transportation, Tongji University, Shanghai 201804, China. (e-mail: \{xuchengkai, yim211, qincheng1, zhanggeyuan, w\_dong, 2111219, guo\_yicheng, hangpeng, sunjian\}@tongji.edu.cn)
}
\thanks{
Jiaqi Liu is with the Department of Computer Science, University of North Carolina at Chapel Hill, United States. (e-mail: jqliu@cs.unc.edu)
}
\thanks{Corresponding author: Peng Hang}}

\maketitle

\begin{abstract}
Autonomous driving is a cornerstone technology for the future of intelligent transportation, where end-to-end learning has emerged as a transformative paradigm that directly maps multimodal sensory inputs to driving actions through unified differentiable models. While offering advantages, the effectiveness of end-to-end autonomous driving (E2E-AD) is ultimately determined by the quality of its training ecosystem. 
This paper provides a comprehensive review of training methods and ecosystem for E2E-AD. We introduce a Data-Strategy-Platform taxonomy that conceptualizes training as an interdependent system. 
The data layer defines what can be learned, the strategy layer governs how learning aligns with driving objectives, and the platform layer supports scalability and continuous evolution. Within this framework, we survey recent advances across data-centric pipelines, learning paradigms, and training infrastructures, and analyze their interplay in shaping model performance, robustness, and deployability.
Finally, we reflect on current limitations and articulate a forward-looking vision that emphasizes a shift from data quantity to data value, from isolated optimization to foundation-driven generalization, and from static training to integrated training-testing loops, aiming toward robust, scalable, and trustworthy autonomous driving systems.
We maintain a continuously updated repository tracking cutting-edge literature and works at \href{https://github.com/Jiaaqiliu/Awesome-Training-Ecosystem-for-E2E-AD}{Our Project Page}.
\end{abstract}


\IEEEpeerreviewmaketitle

\section{Introduction}
\IEEEPARstart{A}{utonomous} driving has emerged as a strategic frontier in intelligent transportation systems, offering prospects for safer mobility, higher traffic efficiency, and new paradigms of urban transport \cite{yan2023learning, feng2021intelligent}.
Traditional research has largely followed a modular pipeline architecture, where perception, prediction, planning, and control are separated into distinct subsystems.
Although this decomposition facilitates interpretability and engineering controllability, modular systems are prone to error accumulation across interfaces and struggle to achieve global optimization in highly interactive and mixed traffic environments \cite{wang2025generative, hu2023planning}.

In response, end-to-end autonomous driving (E2E-AD) has attracted significant research attention.
By mapping multimodal sensory inputs directly to driving actions through unified differentiable models, E2E-AD reduces system coupling and allows joint optimization of perception, decision-making, and control \cite{zhao2024autonomous, le2022survey, chen2024end}.
Such integration has demonstrated strong potential for handling complex interactions and long-horizon reasoning \cite{chib2023recent}.
Moreover, recent progress in large-scale representation learning, generative models, and foundation models further expands the capabilities of E2E-AD, pushing the research frontier from concept demonstrations toward practical deployment.

Despite these structural advantages, E2E-AD still faces challenges in large-scale real-world application, and its development bottleneck lies not only in model design but also in the training process itself.
As models become larger and more capable \cite{cui2024survey}, their effectiveness increasingly depends on whether the training ecosystem can provide: (i) data with sufficient coverage, quality, and governance to capture rare but safety-critical scenarios; (ii) strategies for optimization that can align multi-task objectives and stabilize long-horizon learning; and (iii) platforms with the computational scale and lifecycle management required for continuous model evolution. These three dimensions jointly define the performance boundary of E2E autonomous vehicles.

We clarify the positioning of this review against representative surveys in Table~\ref{tab:survey_positioning}. Existing studies provide valuable reviews of datasets \cite{liu2024survey, wang2024survey, karangwa2023vehicle}, perception modules \cite{zhang2023perception, xiang2023multi}, end-to-end architectures \cite{chib2023recent, chen2024end}, imitation learning \cite{le2022survey}, and foundation models \cite{cui2024survey,zhou2024vision,jiang2025survey}. In contrast, this survey focuses on the training ecosystem and examines how data, learning strategies, and platforms jointly determine training effectiveness.

The main contributions of this survey are as follows:
\begin{itemize}
    \item A Data-Strategy-Platform taxonomy is introduced, which frames E2E-AD training as a system-level challenge rather than an isolated algorithmic problem.
    \item A comprehensive review of recent advances is conducted, encompassing data, paradigms, and infrastructures, with particular emphasis on both the current state of the art and the dynamic interplay among these three dimensions.
    \item Beyond surveying existing work, a future vision for training E2E-AD is proposed, where the shift from data quantity to data value, from task-specific learners to universal foundation models, and from static offline training to continuous training-testing integration is highlighted.
\end{itemize}

\begin{figure*}
    \centering
    \includegraphics[width=\linewidth]{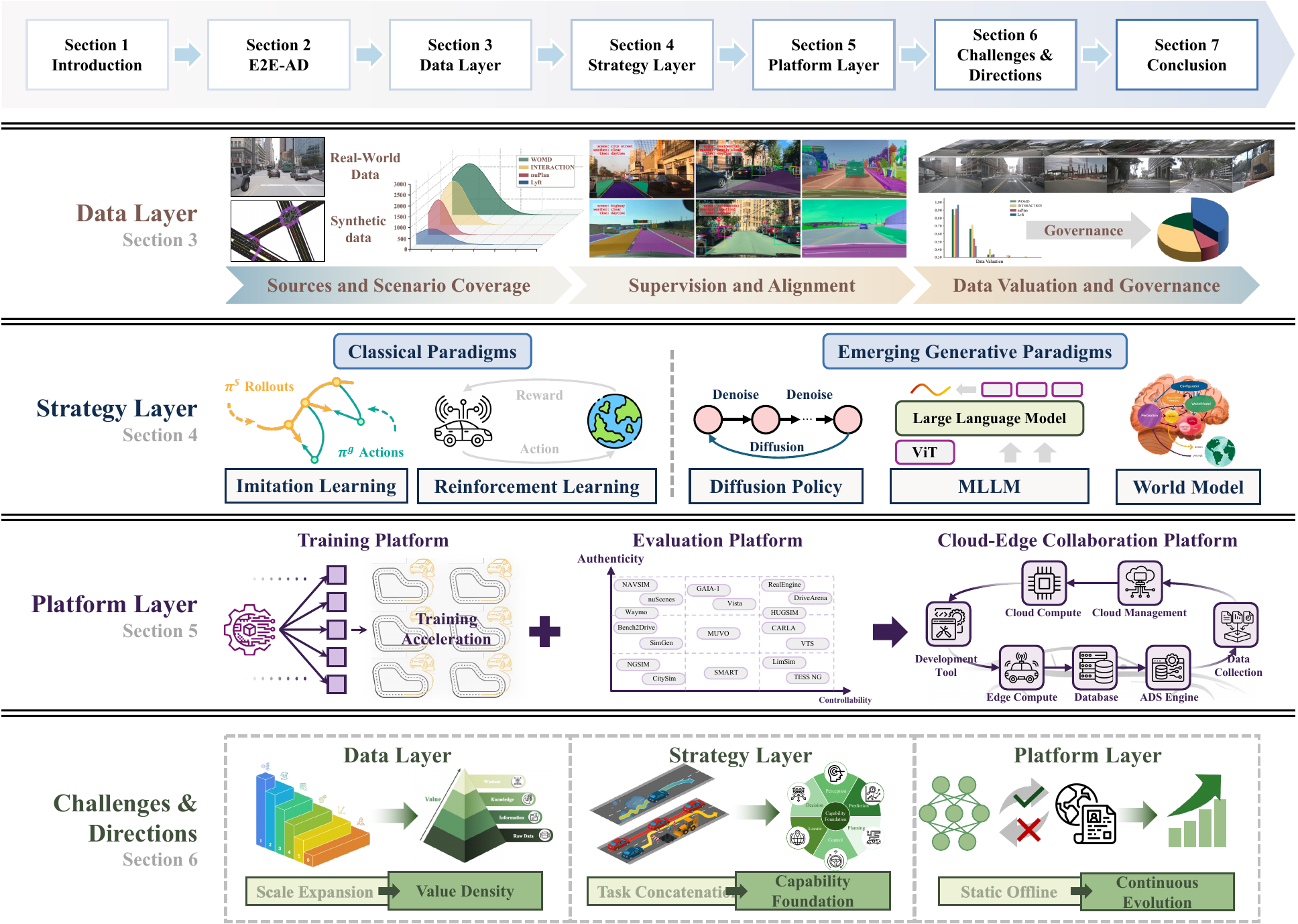}
    \caption{
    Overall structure of the survey. The review is organized around a three-layer training ecosystem.
    (i)\hyperref[Data Layer]{\textit{Data Layer}} illustrates the progression from raw sources to value-driven governance.
    (ii)\hyperref[Strategy Layer]{\textit{Strategy Layer}} contrasts classical paradigms (IL/RL) with emerging generative models (Diffusion, MLLM, World Models).
    (iii)\hyperref[Platform Layer]{\textit{Platform Layer}} shows the infrastructure support from training acceleration to cloud-edge collaboration.
    (iv)\hyperref[Open Challenges And Future Directions]{\textit{Challenges and directions}} highlights the transition toward value density and continuous evolution.}
    \label{fig:overview}
\end{figure*}

The overall logical structure of the review is shown in Fig.~\ref{fig:overview}. The Data Layer (Section \ref{Data Layer}) establishes the learnable information boundary through scenario coverage and governance, which supports the Strategy Layer (Section \ref{Strategy Layer}), where paradigms ranging from imitation learning to world models are reviewed.
Finally, the Platform Layer (Section \ref{Platform Layer}) provides the computational foundation for training and validation. Section \ref{Open Challenges And Future Directions} discusses the cross-layer evolution shown in Fig.~\ref{fig:overview}.

\section{End-to-End Autonomous Driving}
\label{End-to-End Autonomous Driving}

\begin{table}[t]
    \centering
    \caption{Comparison with representative surveys}
    \label{tab:survey_positioning}
    \scriptsize
    \begin{tabular}{l c c c l}
    \toprule
    {\textbf{Survey}} & {\textbf{Data}} & {\textbf{Str.}} & {\textbf{Plat.}} & {\textbf{Main focus}} \\
    \midrule
    {Zhou et al.~\cite{zhou2024vision}} & \XSolidBrush & \Checkmark & \XSolidBrush & MLLMs \\
    {Cui et al.~\cite{cui2024survey}} & \XSolidBrush & \Checkmark & \XSolidBrush & MLLMs \\
    {Chen et al.~\cite{chen2024end}} & \Checkmark & \Checkmark & \XSolidBrush & E2E Architecture \\
    {Zhao et al.~\cite{zhao2025survey}} & \XSolidBrush & \Checkmark & \XSolidBrush & Deep Learning \\
    {Wu et al.~\cite{wu2024recent}} & \XSolidBrush & \Checkmark & \Checkmark & Reinforcement Learning \\
    {Jiang et al.~\cite{jiang2025survey}} & \XSolidBrush & \Checkmark & \XSolidBrush & MLLMs \\
    {Wang et al.~\cite{wang2024survey}} & \Checkmark & \XSolidBrush & \XSolidBrush & Datasets \\
    {Liu et al.~\cite{liu2024survey}} & \Checkmark & \XSolidBrush & \XSolidBrush & Datasets \\
    {\textbf{Ours}} & \Checkmark & \Checkmark & \Checkmark & Training ecosystem \\
    \bottomrule
    \end{tabular}
\end{table}

\subsection{Development and Trends}
Early autonomous driving systems largely relied on modular architectures, which offer interpretability and engineering modularity but suffer from error propagation and limited global optimization in complex mixed traffic.
With the advancement of deep learning and multimodal sensing, E2E-AD weakens information barriers between modules and enables integrated modeling of semantics, interaction, and control by mapping raw sensory inputs directly to driving actions through unified differentiable models\cite{zhao2024autonomous}.
The earliest demonstrations, such as ALVINN and later NVIDIA’s DAVE-2 \cite{pomerleau1988alvinn, bojarski2016end}, showed that neural networks could replace rule-based pipelines with data-driven control policies. While pioneering, these approaches relied on single networks with direct control outputs and struggled to cope with long-tail scenarios or domain transfer.

Subsequent work introduced more structured forms of end-to-end learning, often referred to as modular end-to-end approaches. Instead of discarding intermediate representations, they embedded them within a differentiable framework, allowing perception, prediction, and planning to be optimized under a shared objective. Systems such as MP3, UniAD, and DiffStack \cite{casas2021mp3, hu2023planning, karkus2023diffstack} exemplify this trend to integrate multimodal information and by incorporating differentiable planning modules to improve consistency between perception and control, which alleviates the interpretability issues associated with integrated networks while retaining the benefits of joint learning.

More recently, the field has increasingly embraced generative end-to-end paradigms that capitalize on advances in language models, diffusion models, and related approaches \cite{zhou2024vision, gao2024survey, cui2024survey}. 
These systems move beyond reactive mappings by anticipating long-horizon interactions and synthesizing counterfactual scenarios within latent space.
For example, frameworks such as LMDrive and OpenDriveVLA align vision, language, and action in a shared semantic space, thereby facilitating the natural integration of high-level instructions \cite{shao2024lmdrive, zhou2025opendrivevla}.
Meanwhile, world models such as DriveDreamer and GAIA generate scene rollouts that capture rare or cross-domain conditions \cite{wang2024drivedreamer, zhao2025drivedreamer, hu2023gaia, russell2025gaia}.
These advances mark a transition from narrowly engineered control policies toward models with enhanced capacities for reasoning, generalization, and even imagination.

Alongside architectural evolution, research trends reveal deeper transformations in how E2E-AD is conceptualized.
The data paradigm is moving from scale expansion toward scenario prioritization, with increasing attention to rare, high-risk events and generative supplementation\cite{tian2024tokenize, jiang2025naturalistic, liu2024survey}.
Training objectives are shifting from perception accuracy to consistency with planning and safety, motivating the exploration of imitation learning, reinforcement learning, and generative models\cite{yang2024human, xu2025towards, schrum2024maveric}.
Evaluation protocols are evolving from offline metrics to closed-loop benchmarks such as CARLA and nuPlan, which emphasize safety, comfort, and compliance under continuous deployment cycles\cite{jia2024bench2drive, caesar2021nuplan, zhang2022rethinking, dosovitskiy2017carla}.
These trends illustrate that the progress of E2E-AD is not merely an architectural evolution but a systematic reconfiguration of training logic.

\subsection{Core Training Challenges}
\begin{figure}
    \centering
    \includegraphics[width=0.8\linewidth]{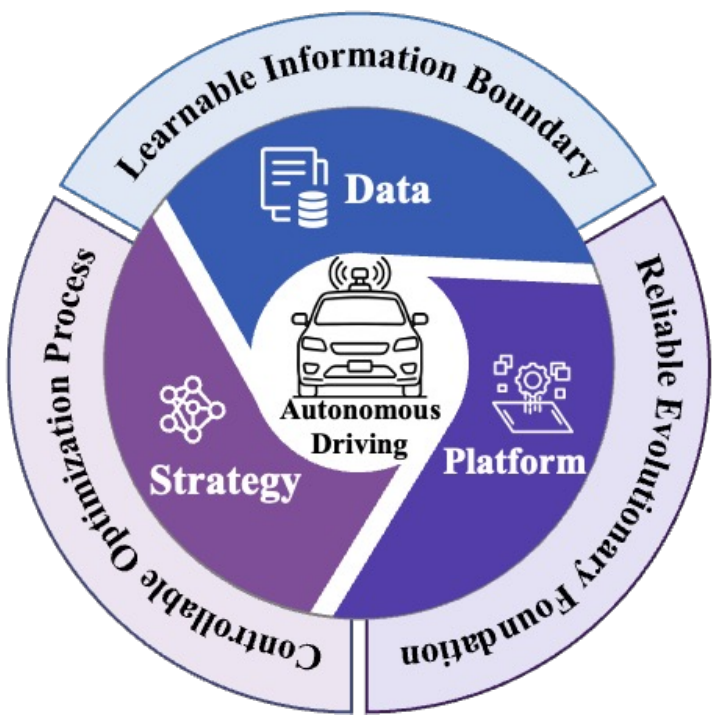}
    \caption{Interrelationship among data, strategy, and platform layers, which illustrates the systemic coupling of the three layers that constitute the E2E-AD training ecosystem. The data layer defines the learnable information boundary, the strategy layer governs the controllable optimization process, and the platform layer provides the reliable evolutionary foundation, which jointly determine the upper bound of autonomous driving capability.}
    \label{fig:interaction}
\end{figure}

As E2E-AD integrates perception, prediction, planning, and control into a unified framework, its success is determined less by any single algorithmic advance and more by the coherence of the training ecosystem. As illustrated in Fig.~\ref{fig:interaction}, three dimensions emerge as decisive. The \textbf{data layer} defines the boundaries of what can be learned, yet the long-tail distribution of road events, the scarcity of high-quality annotations, and discrepancies between real and simulated data severely constrain generalization. The \textbf{strategy layer} governs how models optimize toward driving objectives, but multi-task conflicts, unstable convergence, and the misalignment between offline proxy metrics and online performance frequently undermine reliability. The \textbf{platform layer} provides the engineering foundation, where bottlenecks in data pipelines, distributed training, and continuous adaptation often limit scalability and reproducibility.

These challenges are not independent. Insufficient scenario coverage amplifies optimization instability; fragile optimization increases the burden on computational infrastructure; and platform inefficiencies in turn restrict the feasible scope of data and strategies. Well-known issues such as hallucinations, catastrophic forgetting, or cross-domain degradation often manifest when alignment breaks down across these three layers. Thus, the core challenges of E2E-AD are fundamentally systemic, arising from the need to coordinate data, strategies, and platforms into a closed-loop training process. This interdependence motivates the taxonomy adopted in the remainder of this review, which analyzes training methods along the axes of data, strategy, and platform to provide a unified perspective on the development of E2E-AD.

\section{Data Layer}
\label{Data Layer}

\subsection{Sources and Scenario Coverage}
\label{sources_and_scenario_coverage}

Scenario coverage defines the learnable boundary of E2E-AD. It is not a matter of accumulating more frames, but of exposing the policy to interactions that shape closed-loop safety, robustness, and generalization. A scenario is therefore treated as a temporally bounded traffic interaction conditioned on road geometry, agents, rules, and environment; its value depends on its expected contribution to planning-relevant performance rather than its frequency alone.

Data sources can be organized into three complementary forms.
\textit{Real-world data} from instrumented fleets \cite{huang2018apolloscape, sun2022shift}, shadow-mode operations \cite{silva2022ride}, roadside sensing \cite{zimmer2024tumtraf, zimmer2023tumtraf, xu2023v2v4real, ye2022rope3d, busch2022lumpi}, and crowdsourced vehicles \cite{xia2023automated, longo2017crowd} remain the foundation of E2E-AD training for their naturalistic interactions and region-specific driving habits, yet is costly and inherently stochastic, especially for long-tail and safety-critical events \cite{geiger2012we, wilson2023argoverse, caesar2020nuscenes, mao2021one, sun2020scalability, alibeigi2023zenseact}.
In contrast, \textit{Virtual data} from high-fidelity simulators \cite{dosovitskiy2017carla, qiao2024trafficgamer, guo2025interactive}, 3D reconstructions \cite{chen2025s, zhu2024scene, liao2025learning}, or generative world models \cite{wang2024driving, wang2024drivedreamer, zhao2025drivedreamer, lu2024scenecontrol, xu2025diffscene, wang2025generative} offer controllable expansion of rare or risky conditions but require careful calibration to preserve behavioral realism and prevent distributional drift from real-world dynamics \cite{hendrycks2022scaling, sun2022shift, xu2022v2x, wang2024deepaccident}. Emerging alongside these is \textit{Synthetic data}, produced by generative AI models and typically released after being manually verified.

{In recent years, world models are reshaping the data layer.
They learn generative dynamics that can produce temporally coherent, action- or trajectory-conditioned futures. 
Cosmos-Drive-Dreams, DriveDreamer and GAIA show how such models can synthesize driving scenes, replay counterfactual futures, and connect visual generation with motion-level objectives \cite{wang2024drivedreamer, zhao2025drivedreamer, hu2023gaia, russell2025gaia, ren2025cosmos}. Thus, when used for scenario generation, long-tail augmentation, failure replay, or virtual-real fusion, world models act as data engines; when reused for planning or policy optimization, they enter the Strategy Layer.}

{Effective coverage therefore requires a value-guided hybrid acquisition strategy. Real-world data provides distributional grounding; simulation offers rule-controllable perturbation; and curated synthetic datasets deliver quality-controlled samples. \cite{song2023synthetic, bai2024bridging}}
As the field transitions from quantitative accumulation to qualitative optimization, the landscape of available datasets has shifted accordingly. Table~\ref{tab:datasets_value_scenarios} categorizes representative benchmarks, distinguishing between earlier datasets emphasizing sheer scale (Part A) and recent efforts prioritizing scenario value and rarity (Part B).

\begin{table*}[htbp]
    \centering
    \caption{Representative Datasets on Data Scale and Value}
    \renewcommand{\arraystretch}{1.25}
    \begin{tabular*}{\textwidth}{@{\extracolsep{\fill}} l c c c c c c c c  @{}}
    \toprule
    \textbf{Dataset} & \textbf{Year} & \textbf{Scale} & \textbf{Camera} & \textbf{LiDAR} & \textbf{RaDAR} & \textbf{QA} & \textbf{Type} & \textbf{Data Source} \\
    
    \midrule
    \rowcolor{gray!20}
    \multicolumn{9}{c}{\textbf{(A) Emphasizing Data Scale}} \\
    \midrule
    
    DDD17~\cite{binas2017ddd17} & 2017 & 12 hours & \Checkmark & \XSolidBrush & \XSolidBrush & \XSolidBrush & Real  & Europe  \\
    BDD~\cite{yu2017baidu}	& 2017 & 10000+ km & \Checkmark & \Checkmark &\XSolidBrush & \XSolidBrush & Real & China \\
    DDD20~\cite{hu2020ddd20} & 2020 & 51 hours & \Checkmark & \XSolidBrush & \XSolidBrush & \XSolidBrush & Real & USA  \\
    A2D2~\cite{geyer2020a2d2} & 2020 & 41,277 clips & \Checkmark & \Checkmark &\XSolidBrush & \XSolidBrush & Real & Germany  \\
    Nuscenes~\cite{caesar2020nuscenes} & 2021 & 40K clips & \Checkmark & \Checkmark & \Checkmark & \Checkmark* & Real & USA \& Singapore  \\
    DriveLM~\cite{sima2024drivelm} & 2023 & 188K frames & \Checkmark & \Checkmark & \XSolidBrush & \Checkmark & Virtual \& Real & CARLA \& Nuscenes \\
    DriveMLM~\cite{wang2023drivemlm} & 2023 & 8 maps & \Checkmark & \Checkmark & \XSolidBrush & \Checkmark & Virtual \& Synthetic & CARLA  \\
    CoVLA~\cite{arai2025covla} & 2024 & 6M frames & \Checkmark & \XSolidBrush & \XSolidBrush & \XSolidBrush & Real \& Synthetic & Japan  \\
    OpenScene~\cite{openscene2023} & 2024 & 120+ hours & \Checkmark & \Checkmark &\XSolidBrush & \XSolidBrush & Real & USA \& Singapore   \\
    PhysicalAI~\cite{NvidiaPhysicalAIAutonomousVehiclesDatasets2025} & 2025 & 310,895 clips & \Checkmark & \Checkmark & \Checkmark & \XSolidBrush & Real & Global  \\
    
    \midrule
    \rowcolor{gray!20}
    \multicolumn{9}{c}{\textbf{(B) Emphasizing Data Value}} \\
    \midrule

    Bench2Drive~\cite{jia2024bench2drive} & 2024 & 13,628 clips & \Checkmark & \Checkmark & \XSolidBrush & \Checkmark & Virtual \& Synthetic & CARLA \\
    DriveCoT~\cite{wang2024drivecot} & 2024 & 1,058 scenarios &  \Checkmark & \Checkmark & \XSolidBrush & \Checkmark & Virtual \& Synthetic & CARLA \\
    StyleDrive~\cite{hao2025styledrive} & 2025 & 30K scenarios & \Checkmark & \Checkmark & \XSolidBrush & \Checkmark & Real & OpenScene \\
    AoTA~\cite{guan2025world} & 2025 & 4,800 clips & \Checkmark & \XSolidBrush & \XSolidBrush & \XSolidBrush & Synthetic & BDD100K  \\
    TAD-E2E~\cite{liu2025tad} & 2025 & 1M frames & \Checkmark & \Checkmark & \XSolidBrush & \XSolidBrush & Real & Unknown \\
    WOMD-E2E~\cite{xu2025wod} & 2025 & 4,021 clips & \Checkmark & \Checkmark & \XSolidBrush & \Checkmark* & Real & USA \\
    
    \bottomrule
    \end{tabular*}

    \begin{tablenotes}
    \footnotesize
    \raggedleft
    \item \textsuperscript{*} refers to supplementary data, not the dataset itself.
    \end{tablenotes}
    \label{tab:datasets_value_scenarios}
    
\end{table*}

\subsection{Supervision and Alignment of Labels}
\label{Supervision and Alignment of Labels}

Supervision transforms raw data into learnable signals, enabling E2E-AD systems to distinguish between behaviors that should be learned and those that should be avoided.
As raw sensor data lacks semantic or behavioral grounding, it is annotation that serves as the mechanism for assigning value and meaning to each scenario within the operational design domain (ODD) \cite{liu2024survey}.
Therefore, the goal of supervision is not simply to label data exhaustively, but to produce training signals that are consistent with scenario value and aligned with the system’s planning and safety objectives.

Manual annotation provides high-fidelity supervision by capturing precise object semantics, trajectories, and interaction cues \cite{cheng2022masked, cheng2021per}. However, it is inherently costly and difficult to scale to the volume required by E2E-AD.
Automatic annotation based on pre-trained models, cross-modal fusion, or pseudo-labeling \cite{li2024womd, hirose2025learning, gui2024survey, li2024drivingdiffusion} offers scalability but introduces risks of misalignment, temporal drift, and noise \cite{chen2024end, liu2024survey}. 
In practice, an effective annotation strategy should combine manual annotation for high-value scenarios that are safety-critical or rare and automated or semi-supervised approaches for low-value, high-frequency scenarios where approximate supervision is sufficient.

\begin{figure}
    \centering
    \includegraphics[width=\linewidth]{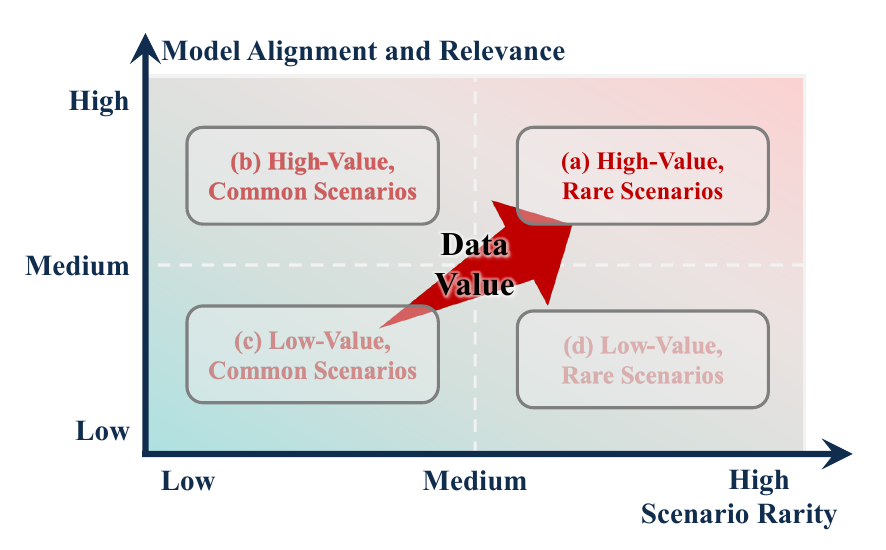}
    \caption{Illustration of classification of scenarios by data value, which divides scenarios into four categories, highlighting the importance of prioritizing high-value scenarios for efficient training.}
    \label{fig:datavalue}
\end{figure}

However, not all labeled data contribute equally to strategy improvement. 
Unlike previous approaches that equated long-tail scenarios with high-value scenarios, this paper argues that scenarios with higher value can bring greater performance gains, especially those that expose decision ambiguities, complex interactions, or boundary conditions of the driving policy, {as depicted in Fig.~\ref{fig:datavalue}.}
As a result, annotation priorities should follow a value-guided hierarchy, with detailed and multimodally aligned labeling for high-value scenarios, while efficient and lower-fidelity supervision for common, low-impact ones.
This approach ensures that annotation resources are concentrated where they most enhance planning stability and safety metrics, rather than where labeling is easiest or most frequent.

\subsection{Data Valuation and Governance}
\label{Data Valuation and Governance}

As the control system of the data layer, data valuation and governance aim to allocate resources toward the most informative and safety-relevant scenarios, thereby accelerating model convergence and improving overall training efficiency\cite{xie2024diversified, li2024boosting, song2022scenario}.
While early industrial practices often pursued data accumulation through sheer volume\cite{NvidiaPhysicalAIAutonomousVehiclesDatasets2025}, the complexity and long-tail characteristics of real-world driving environments suggest that massive low-value samples yield rapidly diminishing marginal returns. Consequently, the paradigm must shift from scale expansion to value density, where the training benefit is determined by the quality of scenario coverage rather than total sample count.

To implement this shift, a quantifiable scenario value system is essential. It summarizes recurring ideas in scenario selection, uncertainty-based mining, and data governance into a simple value-scoring view. We define the value of a scenario $V(s)$ not as a static property, but as its marginal contribution to the model's current capability boundary\cite{yang2025quantitative}. In practice, this can be operationalized through a composite scoring function that integrates risk, uncertainty, and rarity:
\begin{equation}
    V(s) = w_r \cdot R(s) + w_u \cdot U(s | \theta) + w_d \cdot D(s | \mathcal{D}_{train})
\end{equation}
where $R(s)$ quantifies safety-criticality based on physical metrics, $U(s | \theta)$ represents the model's epistemic uncertainty regarding the scenario and $D(s | \mathcal{D}_{train})$ Measures the distributional distance of scenario $s$ from the existing training distribution $\mathcal{D}_{train}$.

Critically, scenario value is dynamic; as the model $\theta$ evolves, previously difficult scenarios ($U(s|\theta) \downarrow$) lose their training utility, while new edge cases gain importance\cite{liu2023mtd, xu2025diffscene}. Effective governance requires a data engine that continuously re-evaluates samples. This involves establishing version trees that map specific data subsets to model checkpoints, ensuring that every performance improvement or degradation can be traced back to the addition or removal of specific high-value scenarios\cite{gheisari2024cappad, wu2023add}. By rigorously managing data lineage and prioritizing resources based on the $V(s)$ metric, the training ecosystem transforms from a passive repository into an active optimization engine, ensuring that computational resources are concentrated where they yield the greatest robustness benefits.

\section{Strategy Layer}
\label{Strategy Layer}

The E2E-AD strategy layer maps multimodal perception to control actions, governing closed-loop stability in mixed traffic.
To transcend the efficiency-robustness trade-offs inherent in Imitation Learning (IL) and Reinforcement Learning (RL), the field is pivoting toward generative paradigms. Approaches such as diffusion policies, multimodal language models, and world models introduce advanced reasoning and adaptability capabilities. As shown in Table~\ref{tab:paradigm_compare}, these paradigms possess distinct, complementary advantages.

\subsection{Imitation Learning}
IL trains an E2E-AD policy directly from expert demonstrations, avoiding explicit reward engineering and yielding strong data efficiency at initialization \cite{hussein2017imitation}. Canonical instances include behavior cloning (BC)\cite{torabi2018behavioral, florence2022implicit, codevilla2019exploring} and distribution-matching variants such as inverse reinforcement learning (IRL)\cite{huang2023conditional, adams2022survey} and generative adversarial imitation learning (GAIL)\cite{ho2016generative, couto2024hierarchical}, which collectively demonstrate that policy-level knowledge can be distilled from human drivers.

Despite their practicality, IL methods face three structural limitations in autonomous driving. First, they exhibit a high dependence on expert data, where demonstrations are costly to collect, often noisy, and may contain suboptimal behaviors \cite{hawke2020urban,le2022survey}. Second, they suffer from covariate shift, as models trained on expert trajectories may encounter unseen states at deployment \cite{chang2021mitigating, seo2024mitigating}, leading to compounding errors and potential loss of control. Third, real-world driving data follow a long-tail distribution, where critical but rare events are underrepresented, leaving IL-trained policies fragile under safety-critical conditions \cite{zhang2025systematic, hallgarten2024can}. These challenges constrain the reliability and generalization of IL in real deployment.

To address these limitations, substantial effort has been devoted to optimizing IL training.
On the data side, augmentation strategies such as viewpoint transformation, trajectory perturbation, and synthetic counterfactual construction enrich the training distribution and mitigate overfitting to narrow expert trajectories \cite{pan2025transfer, ma2025novel, yu2025enduring, wang2024counterfactual, ye2023counterfactual}.
On the algorithmic side, dataset aggregation (DAgger) introduces iterative corrections by allowing the agent to query experts in newly visited states \cite{ross2011reduction}, while SafeDAgger further reduces expert intervention costs by activating corrective supervision only in high-risk conditions \cite{bicer2019sample, prakash2020exploring, ahn2022autonomous}.
Adversarial extensions, such as GAIL and its robust variants, improve generalization by aligning trajectory distributions rather than replicating individual actions \cite{couto2024hierarchical, tang2023personalized}.
More recently, IL has been integrated with generative models.
World models can simulate imagination rollouts of rare or unseen states, providing synthetic replay data to reduce distribution shift without incurring additional safety risks during online collection \cite{popov2024mitigating, cheng2024pluto}.
Together, these techniques transform IL from a purely passive paradigm into an adaptive training mechanism with improved robustness and coverage.

In summary, IL is better viewed as a \textit{policy bootstrapper} rather than a final alignment mechanism in the training ecosystem, which provides a strong prior over normal driving, but isn't able to enforce long-horizon safety under distribution shift by itself. In E2E-AD applications, IL is more suitable for integration with RL for safety alignment and with generative paradigms for scenario imagination and rare-event robustness.

\begin{table*}[t]
\centering
\caption{Comparison of five E2E-AD training paradigms}
\label{tab:paradigm_compare}

\setlength{\tabcolsep}{3pt}
\renewcommand{\arraystretch}{1.20}
\begingroup
\renewcommand\tabularxcolumn[1]{m{#1}}

\begin{tabularx}{\textwidth}{%
  >{\raggedright\arraybackslash}m{1.25cm}
  *{7}{>{\centering\arraybackslash}m{1.00cm}}
  >{\raggedright\arraybackslash}X
  >{\raggedright\arraybackslash}X}
\toprule

\rowcolor{gray!10}
\textbf{Paradigm} &
\thead{\textbf{Multi}\\\textbf{modal}} &
\thead{\textbf{AR}\\\textbf{error}} &
\thead{\textbf{Traj-}\\\textbf{level}} &
\thead{\textbf{RT}\\\textbf{ready}} &
\thead{\textbf{Long-}\\\textbf{horizon}} &
\thead{\textbf{Low}\\\textbf{intera.}} &
\textbf{Interp.} &
\textbf{Best role} &
\textbf{Bottleneck} \\
\hline

IL &
\XSolidBrush & \XSolidBrush & \XSolidBrush & \Checkmark & \XSolidBrush & \Checkmark & \XSolidBrush &
\makecell[l]{%
   Fast policy prior; \\
   Supervised initialization} &
\makecell[l]{%
   Closed-loop drift; \\
   Long-tail blind spots} \\

\rowcolor{gray!10}
RL &
\XSolidBrush & \XSolidBrush & \XSolidBrush & \Checkmark & \Checkmark & \XSolidBrush & \XSolidBrush &
\makecell[l]{%
   Objective-aligned refinement;\\
   Closed-loop optimization} &
\makecell[l]{%
  Reward and exploration risk;\\
  Sim-to-real gap} \\

DP &
\Checkmark & \XSolidBrush & \Checkmark & \XSolidBrush & \XSolidBrush & \Checkmark & \XSolidBrush &
\makecell[l]{%
  Multimodal trajectory proposer;\\
  Timing consistency} &
\makecell[l]{%
  Sampling latency;\\
  Hard-constraint guarantees} \\

\rowcolor{gray!10}
MLLM &
\Checkmark & \Checkmark & \XSolidBrush & \XSolidBrush & \Checkmark & \Checkmark & \Checkmark &
\makecell[l]{%
  Semantic reasoning;\\
  Intent inference} &
\makecell[l]{%
  Hallucination and control gap;\\
  Real-time cost} \\

WM &
\Checkmark & \Checkmark & \Checkmark & \XSolidBrush & \Checkmark & \Checkmark & \XSolidBrush &
\makecell[l]{%
  Imagination-based planning;\\
  Closed-loop generation} &
\makecell[l]{%
  Rollout drift;\\
  Model bias and hallucination} \\
\bottomrule
\end{tabularx}
\vspace{2pt}
\hfill
\begin{minipage}{0.84\textwidth}
\begin{tablenotes}[flushleft]
\footnotesize
\item[] \raggedleft
AR error = autoregressive error accumulation; 
Traj-level = generates trajectories as coherent objects;\\
RT ready = natively compatible with tight closed-loop latency; 
Low intera. = does not require large-scale on-policy interaction.
\end{tablenotes}
\end{minipage}

\endgroup

\end{table*}

\subsection{Reinforcement Learning}
RL optimizes a driving policy through interaction, improving long-horizon returns under an environment-defined feedback signal (MDP/POMDP) \cite{shakya2023reinforcement, matsuo2022deep, liu2023towards}. In E2E-AD, RL’s distinctive value reflects on refining a policy toward long-horizon objectives (safety margins, comfort, compliance) that are difficult to fully encode in supervised labels rather than learning to drive from scratch\cite{wu2024recent, zhao2024survey}.

However, the direct application of RL to E2E-AD encounters four systemic challenges.
First, sample inefficiency remains a central bottleneck, as convergent policies often require millions of interactions, while real-world data collection is costly and unsafe \cite{lee2024ad4rl,lin2024safety,li2024boosting}.
Second, safe exploration is particularly problematic in driving, since unconstrained trial-and-error may lead to collisions, boundary violations, or traffic infractions, which are unacceptable in deployment \cite{gu2023safe, mo2021safe}.
Third, reward misalignment arises because real driving objectives are multi-dimensional which covers safety, comfort, efficiency, compliance and so on, but are often compressed into scalar reward signals, producing biased optimization and divergence between offline metrics and online safety \cite{cai2021decision, xu2026towards}.
Finally, cross-domain generalization is difficult because policies trained in one environment may degrade severely when transferred across cities, weather, or traffic cultures, while sim-to-real gaps further amplify the distribution shift \cite{hong2025effective, akhauri2021improving}.

To mitigate these issues, diverse optimization strategies have been proposed.
For sample efficiency, offline RL leverages large-scale replay datasets to learn policies without costly online exploration, often regularized to prevent extrapolation errors on out-of-distribution actions \cite{diehl2023uncertainty}.
Hybrid approaches combine imitation learning with reinforcement learning, where demonstrations provide fast initialization and RL subsequently fine-tunes policies through interaction \cite{lu2023imitation,zhu2024decoupled,duan2024prompting}.
World models further enhance efficiency by enabling imagination-based rollouts in latent space, thereby generating synthetic trajectories that extend coverage of rare or dangerous scenarios \cite{yan2025streetcrafter,yan2025drivingsphere,li2024think2drive,ni2025recondreamer}.
Regarding safety, constrained RL incorporates explicit constraints such as collision avoidance and lane keeping into the optimization process, while risk-sensitive methods based on Conditional Value at Risk prioritize robustness under tail events \cite{hu2024long, he2023toward, zhang2024cvar, tang2023uncertainty}.
Shielding techniques, such as control barrier functions or model predictive control filters, can override unsafe actions during execution, reducing risk without halting policy improvement \cite{zhang2025control, hou2024safe, gangopadhyay2022safe}.
For reward alignment, preference-based methods such as RLHF align training signals with human judgments of safe and natural driving behaviors, while hierarchical RL decomposes tasks into high-level decision planning and low-level control \cite{zhuang2024hgrl, lee2024imagination, li2024boosting, shang2025drivedpo}, improving interpretability and multi-objective balance.
Finally, generalization is strengthened by domain randomization and style perturbation in simulation, continual learning to preserve knowledge across deployments, and curriculum learning strategies that progressively expose policies to increasingly complex scenarios \cite{yang2025human, anzalone2022end, shi2024task}.

As a result, RL is best positioned as a \textit{policy alignment} and \textit{refinement engine} in the training ecosystem for its ability to enforce long-horizon objectives beyond labels. Therefore, RL is often used in conjunction with world models or to fine-tune other paradigms such as IL and LLM. If used directly for E2E-AD, RL requires high-fidelity evaluation environments and guardrails that bound exploration and reward misspecification.

\begin{figure}
    \centering
    \includegraphics[width=\linewidth]{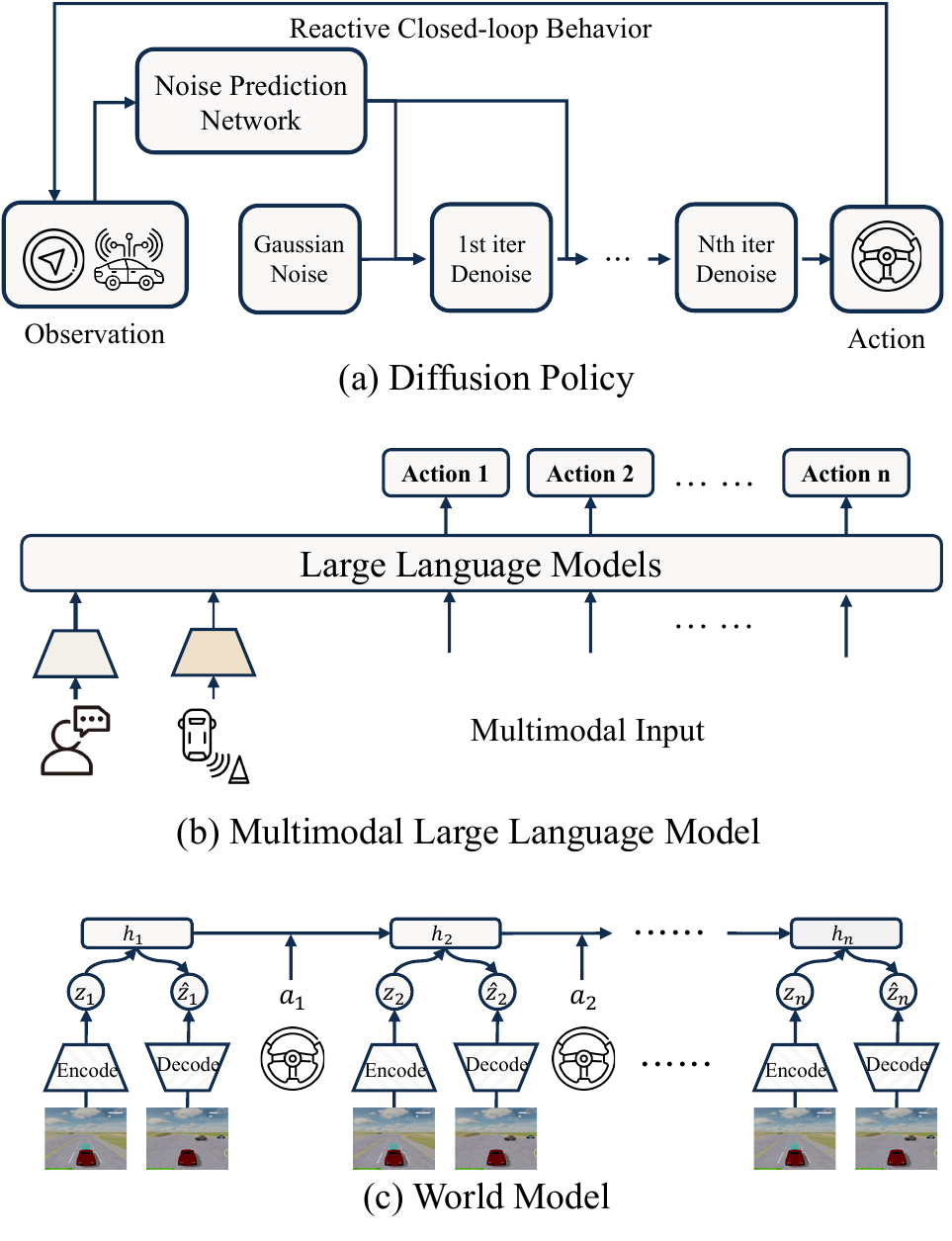}
    \caption{Structure of Emerging Generative Paradigms for Autonomous Driving. This diagram illustrates the integration of generative models in end-to-end autonomous driving systems, covering the application of diffusion policy, multimodal large language model and world model.}
    \label{fig:methodology}
\end{figure}

\subsection{Emerging Generative Paradigms}
Beyond the classical paradigms of IL and RL, recent advances in generative modeling have opened new avenues for E2E-AD. Unlike traditional methods that directly optimize policies over observed data or interactive feedback, generative approaches aim to model richer distributions of actions, semantics, and environments \cite{wang2025generative}, thereby equipping driving agents with capabilities for reasoning, imagination, and adaptation, which signify a transition from purely reactive policy learning toward generative and anticipatory strategies, expanding the scope of training from replication and optimization to reasoning, simulation, and scalable adaptation.

\subsubsection{\textbf{Diffusion Policies}}
Diffusion-based policies represent a generative paradigm for end-to-end autonomous driving, modeling \emph{multimodal} action distributions via a denoising process that refines random noise into structured trajectories or control sequences \cite{ho2020denoising, chi2023diffusion}.
Unlike deterministic mappings in classical policy learning, diffusion policies inherently capture the stochasticity of driving behavior, allowing policies to represent multiple feasible actions under uncertainty \cite{ma2025efficient, peng2025diffusion,li2024learning}.
This property makes them particularly suitable for highly interactive mixed traffic, where safe and robust behavior often depends on anticipating diverse outcomes rather than committing to a single trajectory.

The strengths of diffusion policies lie in their robustness and expressivity.
By modeling the full action distribution rather than a single deterministic path, these methods can naturally account for uncertainty, generate diverse yet plausible driving maneuvers, and avoid mode collapse in decision-making \cite{wu2024dexterous, jiang2025adaptive}.
Such diversity provides resilience in ambiguous or multi-agent interactions, where multiple safe behaviors may coexist. Moreover, the probabilistic nature of diffusion models facilitates uncertainty-aware planning, offering a principled mechanism to balance risk and exploration in safety-critical driving scenarios.

Despite these advantages, diffusion policies also face several critical challenges that recent research is working to address.
The iterative denoising procedure is computationally expensive, elongating training and increasing resource demand\cite{kang2023efficient}.
Correspondingly, efficiency-oriented diffusion variants aim to reduce sampling steps and amortize generation costs. Real-time deployment is further challenged by multi-step sampling latency \cite{chi2023diffusion}, motivating guided or controlled generation that concentrates probability mass on safety-compliant regions and reduces wasted sampling \cite{liao2025diffusiondrive, zheng2025diffusion, zhou2023adaptive}.
Finally, while diffusion is powerful for proposing trajectories, aligning generated outputs with planning objectives and vehicle dynamics is non-trivial, raising concerns about interpretability and integration with safety-critical controllers \cite{liang2024skilldiffuser, hou2024diffusion, wintel2025using}. Scenario-conditioned diffusion and planner-oriented adaptations address this gap by injecting structured context and optimizing generations to satisfy kinematic constraints and downstream safety metrics \cite{jiang2025diffvla, huang2025gen, xu2025knowledge, carvalho2025motion, sun2023conformal}.

In the training ecosystem, diffusion policies are best positioned as a \emph{multimodal proposal and planning component} rather than a standalone E2E controller.
They are most applicable when interaction uncertainty is high and multiple safe futures must be explicitly represented, but they remain constrained by sampling latency, controllability under hard constraints, and the difficulty of certifying safety from a generative distribution.
In practice, these properties make diffusion particularly well-suited to the role of a trajectory-level planner that converts high-level decisions into diverse and feasible candidates.
Moreover, diffusion policies are often most effective in hybrid training pipelines, serving to preserve multimodality and uncertainty awareness in other paradigms.

\subsubsection{\textbf{Multimodal Large Language Models}}
Multimodal large language models (MLLMs), including vision-language-model (VLM) and vision–language–action (VLA) systems, represent an emerging direction for training E2E-AD policies, whose underlying principle is to leverage the grounding capabilities of large-scale language and vision–language models to connect raw perception inputs with natural language instructions and low-level vehicle actions \cite{xie2026latentvla,jiang2025survey,zhou2024vision}.
By aligning sensory observations with symbolic or linguistic abstractions, these models aim to bridge high-level reasoning with control execution, thereby enabling autonomous vehicles to understand and respond not only to the physical environment but also to human commands expressed in natural language.

The key strengths of MLLMs lie in their generalizability and interpretability \cite{yin2024survey,ma2024survey, tian2024drivevlm}.
Owing to large-scale pretraining across diverse multimodal corpora, these models exhibit strong transferability to new scenarios with minimal task-specific data, a property especially valuable for cross-domain driving deployment \cite{dong2025towards, zhou2025opendrivevla, pan2024vlp}.
Furthermore, their ability to process natural language enables command reasoning, which allows human operators to guide vehicle behavior intuitively.
The integration of linguistic representations also provides a degree of transparency and interpretability absent in purely black-box policies, while their modular pretraining structure supports flexible adaptation to varied downstream driving tasks \cite{xing2024autotrust, ding2025hint}.

Nonetheless, significant challenges constrain their practical deployment in autonomous driving.
First, maintaining temporal consistency is difficult, as LLM-driven policies must integrate sequential sensor data and reason over long horizons, whereas most pretrained models are optimized for static image–text pairs \cite{jiang2025diffvla, li2025recogdrive}.
Second, low-level grounding remains limited, as translating high-level linguistic reasoning into precise, real-time control commands requires robust alignment across abstraction layers \cite{long2024vlm, fu2025orion}.
Third, safety and reliability are unresolved concerns, since language models may hallucinate, misinterpret ambiguous instructions, or generate unsafe control suggestions \cite{qu2025vl, zhang2025evaluation, xie2025vlms}.
Finally, issues of real-time performance pose engineering hurdles, as the inference latency of large multimodal architectures is often incompatible with the stringent time constraints of closed-loop driving  \cite{jiang2025structured, liu2025dsdrive, baumann2025enhancing, hegde2025distilling}.

To mitigate these limitations, researchers have proposed specialized training methodologies.
Supervised pretraining on paired perception–language–action datasets improves grounding between natural language and driving trajectories \cite{arai2025covla, marcu2024lingoqa, chi2025impromptu,li2024womd}.
Instruction tuning further enhances the ability to follow human commands in structured traffic contexts \cite{zhang2024wisead, xu2025towards, sheng2025curricuvlm}.
Beyond supervised alignment, RLHF and its variants provide preference-driven optimization, allowing policies to align with human judgments of safety, comfort, and naturalistic behavior \cite{zhou2025autovla}.
More recently, chain-of-thought and visual chain-of-thought reasoning have been introduced to enhance interpretability and robustness, by decomposing multimodal inputs into explicit intermediate steps that link perception, linguistic reasoning, and action selection \cite{zeng2025futuresightdrive, qian2025agentthink,liao2025cot}.
This not only improves transparency in decision-making but also reduces error propagation by anchoring reasoning in visually grounded cues.

In the training ecosystem, MLLMs are best interpreted as \emph{semantic decision and coordination modules}, which are effective at mapping complex perception into structured intents, rules, and explanations.
However, due to temporal inconsistency and the risk of hallucinated outputs, deploying MLLMs directly without safeguards remains difficult.
Practically, MLLMs complement other paradigms by providing verifiable intermediate representations that condition downstream planners and controllers, such as diffusion planners or IL/RL policies.

\subsubsection{\textbf{World Models}}
World models (WMs) represent a generative paradigm that aims to endow E2E-AD systems with the ability to learn compact latent representations of environment dynamics and to use these representations for “imagination”-based planning \cite{tu2025role, xie20252d, long2025survey}.
Instead of directly mapping sensory observations to actions, world models first construct a predictive latent space that encodes the spatiotemporal evolution of the driving environment.
Policies can then be optimized within this latent space by simulating counterfactual trajectories or multi-agent interactions, thereby reducing reliance on costly real-world sampling \cite{feng2025survey, friston2021world, guan2024world}.
This principle fundamentally shifts the training process from purely reactive behavior learning toward proactive reasoning over possible futures.

The primary strengths of world models lie in sample efficiency and scenario generalization.
By replacing expensive environment interactions with latent rollouts, world models dramatically reduce the need for real-world data collection while still supporting long-horizon policy optimization \cite{lin2020improving, sekar2020planning}.
Furthermore, their capacity to generate counterfactual scenarios allows the model to explore safety-critical events that may be scarce in naturalistic datasets but crucial for robust deployment \cite{ding2025understanding}.
As a result, world models are well-suited for addressing the long-tail problem in E2E-AD.

Nevertheless, significant challenges hinder their practical application.
A central issue is the fidelity of generated rollouts, as imperfect latent predictions can diverge from real-world dynamics, undermining policy reliability.
Hallucination risks, where the model produces implausible or inconsistent environment states, pose safety concerns when imagination outputs are used for policy training \cite{gao2024vista, gu2024dome, vafa2024evaluating, zhou2025hermes}.
Moreover, ensuring stability over long horizons remains difficult, as small errors in latent prediction often accumulate, leading to compounding deviations that degrade policy performance \cite{kwon2025toward}.

Recent advances in training strategies seek to address these challenges.
Predictive modeling methods focus on learning latent state transitions that preserve physical plausibility and semantic consistency, often through self-supervised or multi-task objectives \cite{wang2024drivedreamer, wang2024driving, min2024driveworld}.
Imagination-augmented reinforcement and imitation learning integrate world model rollouts into standard policy optimization, expanding the range of training scenarios without requiring direct real-world exploration \cite{li2024imitation, li2024think2drive, yang2025raw2drive}.
Finally, closed-loop adaptation connects imagined experiences with real-world feedback, continuously refining the latent dynamics model to align imagination with observed behavior. These approaches collectively aim to transform world models from auxiliary simulators into central engines of policy training \cite{zheng2024doe, yang2024drivearena}.
   
In the training ecosystem, world models are best positioned as \emph{imagination and evaluation backbones}. 
{As clarified in Section~\ref{sources_and_scenario_coverage}, their first role is to serve as data engines that generate, replay, and diversify rare or high-value scenarios. Recent works further show that unifying visual scene generation with motion representation can make world-model representations transferable from generation to planning \cite{zhang2025epona, gui2026bridging,yang2026worldrft,wang2025adawm,zheng2025world4drive}. However, their effectiveness is ultimately bounded by fidelity, especially with respect to physical rules, causal interaction, and motion consistency rather than visual appearance alone.}
However, their effectiveness is ultimately bounded by fidelity, especially in terms of physical rules rather than optical visual.
It can serve as a joint optimization tool for other paradigms, provide more valuable data for IL, LLM, and other paradigms, and provide closed-loop simulation environment for RL training or evaluation.

\section{Platform Layer}
\label{Platform Layer}

\subsection{Distributed and Parallel Training Platform}
For large-scale end-to-end models, the memory and throughput requirements of training exceed the capacity of a single computing device. Consequently, distributed and parallel training has become a central issue in the infrastructure layer. The design objectives of such systems can be summarized in three dimensions: first, scalability, maintaining near-linear throughput improvement as device scale grows; second, efficiency, achieving efficient overlap and load balancing between communication and computation; third, stability and reproducibility, ensuring numerical consistency and process traceability during long-cycle training. Around these objectives, five complementary parallel paradigms have been developed, which are often combined in hybrid forms in engineering practice.

\subsubsection{\textbf{Data Parallelism (DP)}} 
DP is able to replicate full models across devices, partition training data into separate forward and backward computation, and synchronize gradients via all-reduce \cite{subhlok1993exploiting, sergeev2018horovod, cortes2021object}.
It offers ease of implementation, insensitivity to model architecture, and near-linear throughput scaling. However, its drawbacks include high memory overhead for the reason that parameters and optimizer states must be stored on each device and sensitivity to network bandwidth during gradient synchronization.
Gradient accumulation, compression, and delayed synchronization are usually used to alleviate bandwidth pressure in engineering practices \cite{narayanan2021memory, narayanan2021efficient}.

\subsubsection{\textbf{Tensor Parallelism (TP)}}
TP partitions models at the operator or weight level, distributing large-scale matrix multiplications and convolutions across devices\cite{wang2022tesseract}.
Each device stores only parameter subsets, reducing memory pressure and enabling training at ultra-large parameter scales, whose main cost is fine-grained cross-device communication, where mismatched partitioning and topology can amplify overhead. Efficient implementation requires fusion and scheduling at the operator level to maximize overlap with computation.

\subsubsection{\textbf{Pipeline Parallelism (PP)}}
PP divides models into stages by network depth, enabling micro-batches to flow through the pipeline across stages, which can reduce peak memory via staged storage and checkpointing and improve utilization in deep networks \cite{sun2024adapipe}.
Its effectiveness, however, is constrained by idle ‘bubble’ time and load imbalance, which can be alleviated by finer-grained stage partitioning, careful tuning of micro-batch scheduling, and applying recomputation at stage boundaries to further reduce activation storage \cite{kim2023bpipe, liu2023hanayo}.

\subsubsection{\textbf{Sequence Parallelism (SP)}}
SP distributes activations and contexts along time or token dimensions, distributing memory and communication costs of long sequence training across devices\cite{li2021sequence}.
It is especially effective for Transformers, extending trainable context length while reducing long sequence memory peaks. Typical implementations, such as DeepSpeed Ulysses \cite{jacobs2023deepspeed}, use all-to-all communication to partition attention computation. SP integrates well with DP, TP, and PP, though boundary exchanges introduce additional communication. For the excellent performance optimization, the DeepSpeed architecture is widely used for most existing model pre-training or fine-tuning \cite{miao2023towards, zheng2024llamafactory, zhang2024found, cai2025survey}.

\subsubsection{\textbf{Expert Parallelism (EP)}}
Mixture-of-experts models expand effective parameter scale without proportionally increasing memory and computation by introducing multiple experts in feedforward layers and performing sparse routing on input tokens, activating only a few experts for computation. Expert parallelism distributes different experts across multiple devices, routing tokens to corresponding experts in all-to-all fashion during the forward pass, then aggregating after computation. 
Representative systems like Google's GShard~\cite{lepikhin2020gshard} replace some layers in models with MoE layers, distribute different experts across different devices, and use routers for cross-device communication, achieving parameter sparsification in ultra-large-scale models. Microsoft's DeepSpeed MoE~\cite{rajbhandari2022deepspeed}, based on the DeepSpeed architecture and combining ZeRO optimizer~\cite{rajbhandari2020zero} with All-to-All communication optimization, significantly reduces memory consumption, demonstrating feasible paths for large-scale sparse models and significantly reducing communication and latency costs during training and inference.

\subsubsection{\textbf{Hybrid Parallelism and Strategy Selection}}
In practice, employing hybrid parallel strategies is essential for training large-scale E2E-AD models. While DP and TP typically form the foundation for parameter scaling, PP and SP are increasingly critical for optimizing deep architectures and long-context multimodal streams. Additionally, EP offers sparsification-based scaling to expand model capacity within feasible budgets. The essence of these distributed frameworks lies in communication-aware partitioning that balances memory, computation, and bandwidth pressure across the cluster.

This evolution in parallel strategies is directly driven by the escalating hardware demands of modern algorithms. As detailed in Table \ref{tab:compute_cost}, there is a clear trend toward exponential resource consumption, which is transitioning from classical Imitation Learning models, which were manageable on single-node configurations, to Generative paradigms that necessitate massive multi-node clusters to handle immense parameter counts. This non-linear growth not only validates the necessity for advanced hybrid parallelism but also highlights a widening compute gap between training infrastructure and energy-constrained vehicle edge devices. Consequently, future platforms must extend beyond scalable training to include robust pipelines for model distillation and quantization, ensuring that large-scale foundation models remain deployable in real-world environments.

\begin{table}
    \centering
    \caption{Comparison of Computational Requirements for Representative E2E-AD}
    \label{tab:compute_cost}
    \begin{tabular}{l l c l}
        \toprule
        \textbf{Method} & \textbf{Year} & \textbf{Paradigm} & \makecell[l]{\textbf{Training} \\ \textbf{Hardware}*} \\
        \midrule
        InterFuser\cite{shao2023safety} & CoRL 2022 & IL & 8$\times$V100 \\
        UniAD\cite{hu2023planning} & CVPR 2023 & IL & 16$\times$A100 \\
        DiffusionDrive\cite{liao2025diffusiondrive}  & CVPR 2025  & DP & 8$\times$RTX4090   \\
        AutoVLA\cite{zhou2025autovla} & NeurIPS 2025 & MLLM  & 8$\times$L40   \\
        ORION\cite{fu2025orion}   & ICCV 2025  & MLLM  & 32$\times$A800  \\
        FSDrive\cite{zeng2025futuresightdrive} & NeurIPS 2025   & MLLM  & 32$\times$A6000 \\
        RS2E\cite{liu2025reinforced}  & TPAMI 2026 & \makecell[c]{IL+RL\\+DP}   & 8$\times$A100   \\
        \bottomrule
    \end{tabular}
    \begin{tablenotes}
    \footnotesize
    \raggedleft
    \item \textsuperscript{*} reported experimental hardware, not minimum requirements.
    \end{tablenotes}
\end{table}

\subsection{Testing and Evaluation Platform}

As an essential component of the E2E-AD training system, the testing platform is also responsible for the performance, safety, and reliability of models in real-world or simulated environments \cite{chen2023milestones, tang2023survey, lou2022testing}.

\subsubsection{\textbf{Simulation-Based Testing and Closed-Loop Validation}} 
Simulation-based testing plays a pivotal role in evaluating E2E-AD models in a controlled, repeatable environment.
High-fidelity simulators such as CARLA, LGSVL, and NVIDIA DRIVE Sim allow models to be tested across a wide range of predefined scenarios, including edge cases and rare occurrences that would be costly or dangerous to replicate in real-world trials \cite{wang2022adept,dosovitskiy2017carla,rong2020lgsvl, NVIDIAAutonomousVehicleSimulation,li2022metadrive, jia2024bench2drive}.
Publicly available descriptions indicate that Waymo emphasizes simulation-first validation, using its simulation platform \cite{SimulationCityIntroducing} to replicate billions of virtual miles prior to limited on-road deployment.
These platforms provide virtual environments that can accurately model road conditions, traffic patterns, and weather variations, ensuring comprehensive coverage of possible driving situations.
However, simulation-to-real transfer remains a critical challenge because differences between simulated and real-world environments can lead to discrepancies in model performance, especially when dealing with rare events \cite{zheng2023opencda, hu2023simulation}.

\subsubsection{\textbf{Real-World Testing and Real-Time Evaluation}}
While simulation provides valuable insights, real-world testing is indispensable for validating E2E-AD systems under actual driving conditions \cite{zhou2024would, rossolini2023real}.
On-road testing exposes models to dynamic, unpredictable environments, including complex human behaviors, varying traffic conditions, and environmental factors such as weather or road surface variations \cite{lee2024study}. 
The challenge in real-world testing is to ensure safe operation of the vehicle while evaluating the model's adaptability to real-time data streams and decision-making. 
Closed-loop real-world testing integrates real-time sensor data with the model’s decision-making process, allowing for feedback on how well the model adapts to unforeseen situations.
A robust real-world testing platform must incorporate monitoring and safety mechanisms to mitigate risks associated with algorithm instability, though these safeguards may sometimes mask the model's true performance. Therefore, hybrid virtual-real testing platforms may offer a promising solution for balancing safety with accurate performance evaluation \cite{shoukat2024autonomous, hu2023simulation, argui2024advancements, xu2025tell}.

Based on publicly available descriptions, industrial E2E-AD platforms appear to differ along two broad tendencies, although the internal details of proprietary systems are not fully observable.
A verification-driven, simulation-centric regime, represented by Waymo, treats safety as a pre-deployment property that must be demonstrated through repeatable, scenario-addressable evidence; accordingly, it prioritizes high-fidelity simulators, rule-based scenario generation, and deterministic regression pipelines that support reproducibility, interpretability, and regulatory traceability.
In contrast, an experience-driven, fleet-centric regime, represented by Tesla, treats safety and generalization as emergent properties that improve through large-scale exposure to real operating conditions; it therefore prioritizes event-triggered data capture, rapid data-to-training conversion, and continuous deployment mechanisms that shorten the discovery-to-fix cycle for long-tail failures\cite{TeslasShadowTesting}.
These two concepts are not mutually exclusive, and emerging companies such as XPeng and Huawei may combine elements of both.

\subsubsection{\textbf{Testing Metrics and Performance Validation}}
The efficacy of testing platforms is fundamentally determined by the evaluation metrics used to measure the model’s performance \cite{jia2024bench2drive, caesar2021nuplan}.
Traditional metrics, such as accuracy in object detection or classification, are insufficient to fully capture the complexity of autonomous driving tasks \cite{sharath2021literature}.
Instead, testing platforms must incorporate a broader range of metrics that assess safety, comfort, and efficiency.
Key performance indicators include collision rates, the ability to perform emergency maneuvers, and compliance with traffic laws.
Additionally, metrics such as acceleration smoothness that relate to user comfort are also critical in evaluating the model's effectiveness in providing a human-friendly driving experience \cite{inoue2024nuscenes, li2024ego, li2023toward, xu2022safebench}.

The main challenges in testing platforms revolve around scalability, realism, and safety.
As E2E-AD models grow in complexity, testing platforms must scale accordingly, handling increasing amounts of data and higher computational requirements.
However, most existing tests discover problems through massive testing, which often wastes excessive resources. How to accelerate testing and training by leveraging concepts such as scenario value may be an important research topic for the future.

\begin{figure}
    \centering
    \includegraphics[width=\linewidth]{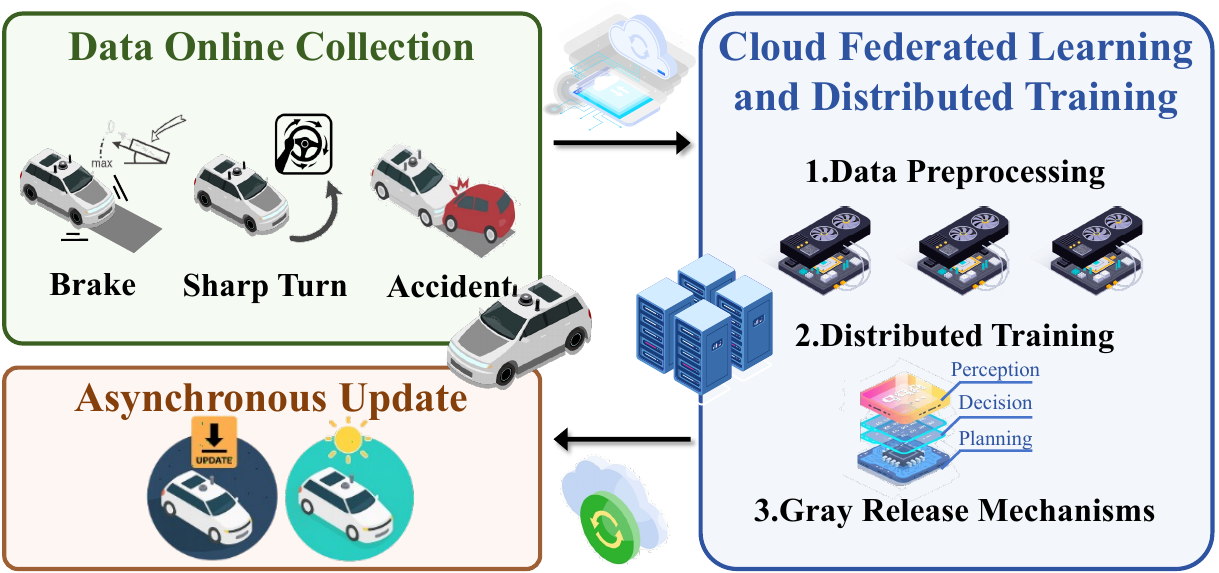}
    \caption{Continuous autonomous driving training framework based on end–cloud collaboration, which illustrates the closed-loop process of continuous model evolution. Vehicle-side data collection captures high-value driving events, which are uploaded for cloud-based federated and distributed training. Through asynchronous updates, models are incrementally improved and redeployed, achieving sustainable and adaptive evolution.}
    \label{fig:collabration}
\end{figure}

\subsection{Cloud–Edge Collaboration and Continuous Training Platform}
As autonomous driving systems move from research and development to actual deployment, models face increasingly complex environmental dynamics and regional differences \cite{li2024platform}.
Relying solely on offline training can no longer meet long-term stability and cross-domain adaptability requirements.
{Public information suggests that XPeng and Huawei ADS \cite{XPENGOfficialWebsite} have explored cloud-edge collaboration frameworks related to Fig.~\ref{fig:collabration}, including event-driven collection, staged updates, and cloud-supported training pipelines.}
Cloud-edge collaboration and continuous training thus become key paths for improving system evolution capability \cite{zhang2022rethinking, verwimp2023clad, li2024data, ge2024life}.
Their core objective is to establish closed-loop pathways between vehicle-side and cloud-side: the vehicle side focuses on event-triggered data collection and lightweight processing, while the cloud side provides large-scale computing support, enabling asynchronous incremental model updates and cross-regional adaptation.

Thus, cloud-edge collaboration not only achieves decoupling and coordination between training and deployment but also provides dynamic adaptation and continuous optimization capabilities for autonomous driving models, serving as core technical guarantees for system large-scale deployment.

\subsubsection{\textbf{Online Collection and Edge Preprocessing}}
After deployment, vehicle-side systems should use event-driven collection strategies as core, prioritizing the recording of samples with high marginal value for safety and robustness, such as emergency braking, sharp turns, close encounters, abnormal yielding, and perception divergence. Vehicle sides complete preliminary quality filtering and redundancy removal, including time synchronization correction, basic calibration verification, and compression encoding, performing lightweight anonymization and desensitization processing locally as much as possible, then uploading in grades according to bandwidth and priority. Compared to traditional offline collection, this mechanism continuously supplements long-tail scenarios at low additional cost and provides high-frequency fresh samples during regional migration and seasonal changes.

\subsubsection{\textbf{Asynchronous Updates and Model Governance}}
Continuous training needs to break hard constraints of synchronous cycles, forming asynchronous mechanisms where data flow, training flow, and deployment flow advance in parallel. After data enters training platforms, incremental training jobs are automatically triggered, producing candidate models and registering them in model repositories. Each candidate version must be bound to clear data versions, configurations, and checkpoints, completing offline metrics and replay set evaluation before entering online validation. To shorten improvement realization time, stable job orchestration and resource scheduling strategies should be established, enabling deep overlap of decoding, augmentation, communication, and computation to improve throughput and timeliness. Meanwhile, simple and clear promotion criteria should be set to ensure update rhythms are predictable and results interpretable.

\subsubsection{\textbf{Gray Update and Rollback Mechanisms}}
New models should not go fully online at once. Gray release follows principles of small scale and controllable scope, first enabling new versions in limited fleets or specific regions, continuously observing safety-related metrics, interaction quality metrics, and system loads, then gradually expanding coverage in phases. To prevent occasional mismatches from evolving into systematic risks, one-click rollback channels must be established: when degradation alerts are triggered or preset thresholds are exceeded, immediately switch to the previous stable version and automatically archive related data to support subsequent localization and review. The coordination of gray release and rollback makes trial and error controllable, enables rapid damage control for failures, while providing high-value failure samples for continuous training.

Industrial efforts toward scalable testing and continuous improvement have increasingly shaped the development of E2E-AD. Publicly reported practices from companies such as Waymo, Tesla, Cruise, Baidu Apollo, XPeng, and Huawei suggest different philosophies regarding generalization, safety validation, and update efficiency. To illustrate this landscape, Table~\ref{tab:industrial_testing} summarizes representative approaches based on public sources; the entries should be read as indicative patterns rather than exhaustive descriptions of proprietary systems.

\begin{table*}[ht]
    \centering
    \caption{Industrial Practices in Autonomous Driving}
    \renewcommand{\arraystretch}{1.25}
    \begin{tabularx}{\textwidth}{l X X X X X}
    \toprule
    \rowcolor{gray!10}
        \textbf{Company} & \textbf{Learning Mechanism} & \textbf{Testing Approach} & \textbf{Key Focus} & \textbf{Advantages} & \textbf{Challenges} \\
    \midrule
    
        Waymo & 
        Continuous retraining with real-world validation &
        Simulation-first using Carcraft platform &
        High-fidelity, rule-based scenario generation for safety validation &
        High reproducibility and interpretability; rigorous safety validation &
        High scalability costs; limited adaptability to real-world conditions \\
    \rowcolor{gray!10}
        Tesla &
        Continuous fleet learning with online retraining &
        Real-world testing through fleet data collection &
        Event-triggered data collection; large-scale fleet feedback; automatic data annotation &
        Scalable exposure to real-world conditions; rapid iteration of training &
        Weak reproducibility; regulatory challenges regarding data privacy and safety \\
     
        Cruise &
        Incremental city-level training updates with controlled testing zones &
        Hybrid real-world and simulation testing &
        City-specific training and safety validation &
        Robust version control; effective safety monitoring within urban environments &
        Limited scalability across regions; expensive infrastructure for testing and validation \\
    \rowcolor{gray!10}
        Baidu Apollo &
        Periodic retraining with continuous deployment in open-road pilots &
        Hybrid real-world and simulation testing &
        Government-integrated data governance and safety compliance &
        Strong regulatory compliance; consistent model reproducibility &
        Slower feedback adaptation; challenges in handling diverse driving environments \\
     
        XPeng &
        Federated learning and asynchronous updates for cross-regional adaptation &
        Cloud-edge collaborative training and testing &
        Cross-region data federation for model adaptation and deployment &
        Efficient cross-domain updates; real-time adaptability to regional differences &
        High infrastructure costs; limited transparency in testing methodologies \\
    \rowcolor{gray!10}
        Huawei ADS &
        Continuous learning with gray release and rollback mechanisms &
        Cloud-edge collaborative training and testing &
        Stable asynchronous updates and safety-focused rollback strategies &
        Reliable model adaptation with real-time updates; controllable risk management &
        Limited openness and transparency; high computational cost for training\\
    \bottomrule
    \end{tabularx}
    \label{tab:industrial_testing}
\end{table*}

\section{Open Challenges And Future Directions}
\label{Open Challenges And Future Directions}

E2E-AD is currently transitioning from scientific exploration to large-scale engineering deployment.
To address the constraints of application, the training ecosystem should evolve from isolated algorithmic optimizations to systemic synergy.
Future developments will synchronize across three axes: increasing data value density, establishing foundation model capabilities, and enabling continuous platform evolution, which can reinforce each other to define the next generation of robust and scalable autonomous systems.

\begin{figure*}
    \centering
    \includegraphics[width=\linewidth]{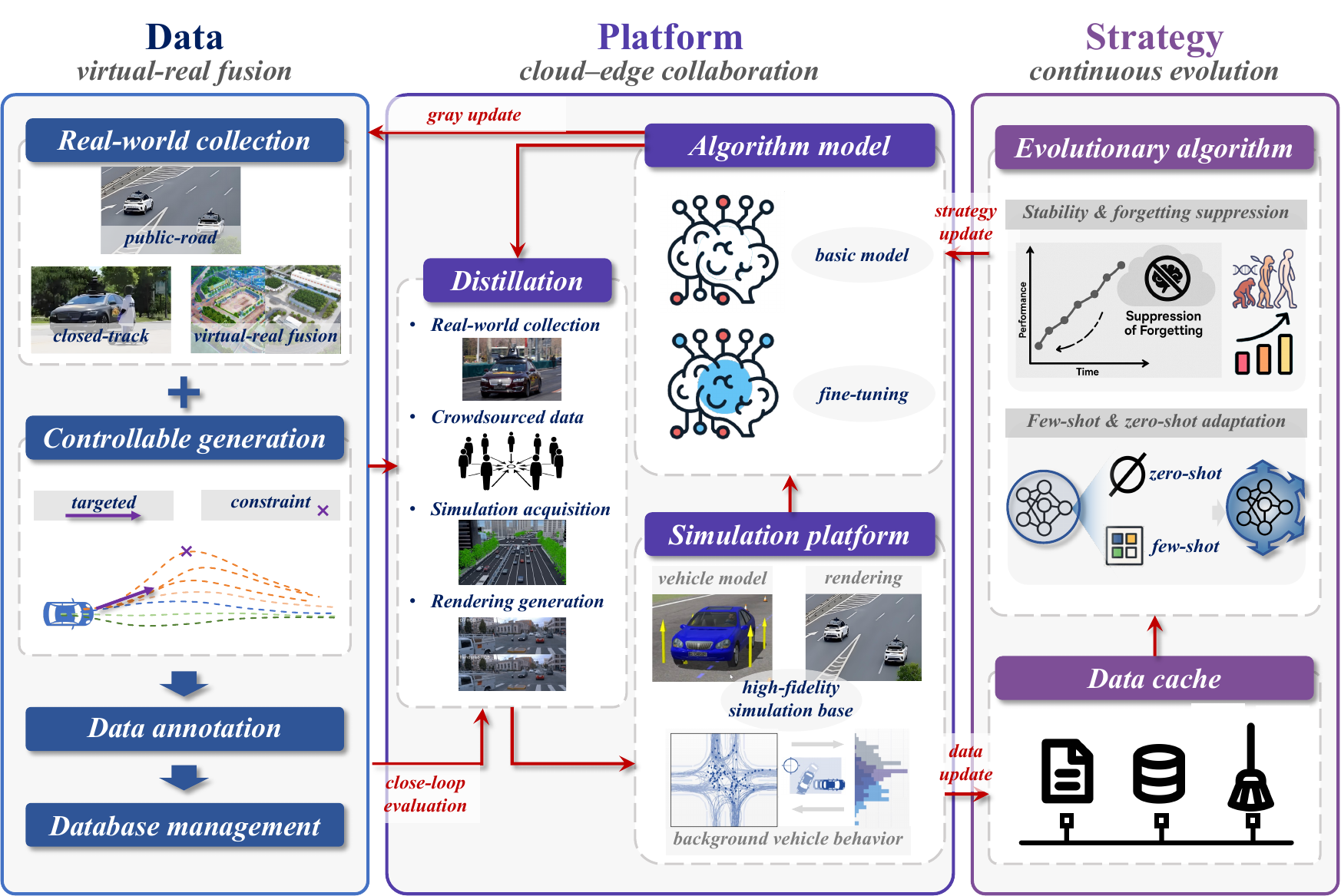}
    \caption{
    Schematic diagram of the Data-Strategy-Platform coupled training system, demonstrating the closed-loop cycle of continuous evolution.
    \textbf{(Left) Data Virtual-Real Fusion}: Real-world failures trigger controllable generation to create targeted training sets.
    \textbf{(Middle) Cloud-Edge Collaboration}: The Platform layer manages the distillation of heavy models into deployable units, facilitating gray updates.
    \textbf{(Right) Strategy Evolution}: The Strategy layer employs evolutionary algorithms to adapt to new scenarios while suppressing catastrophic forgetting.
    The red arrows indicate the critical feedback loop from online deployment back to offline optimization.
    }
    \label{fig:training_testing_integration}
\end{figure*}

\subsection{Scenario Prioritization and Data Engine: From ``Data Volume'' to ``Value Density''}
Building on the value-driven governance defined in Section~\ref{Data Layer}, the primary engineering challenge remains the construction of automated value scoring systems that can operate at scale. While the necessity of "value density" is clear, implementing it requires overcoming significant bottlenecks in current data infrastructures.

First, collection processes rely on fixed routes or random strategies, with insufficient coverage of key scenarios and redundant samples occupying substantial resources. Additionally, annotation processes are inefficient, especially in multimodal data processing where spatiotemporal alignment problems remain unsolved, and lack of hierarchical management leads to high-cost annotation consumption on low-value samples. Finally, governance processes are fragmented, with incomplete version records and difficult traceability of additions and removals, thus limiting reproducibility of model iteration. To achieve continuous progress, systematic breakthroughs are urgently needed in the following aspects:

\begin{itemize}
    \item \textbf{Construct scenario value systems.} Establish quantifiable and operational scenario value scoring systems that comprehensively consider risk, rarity, learnability, and coverage gaps to form unified priority lists. Implement hierarchical treatment and resource allocation based on thresholds, making collection and annotation focus on scenarios with the highest marginal returns.
    \item \textbf{Implement event collection and controllable synthesis.} In event-triggered collection, rely on vehicle-side shadow operation around trigger conditions such as emergency braking, sharp turns, and close encounters to prioritize recording ``near-miss events'' and capture critical decision samples from human driving. On the synthesis side, maintain semantic consistency and physical credibility, using programmatic generation and physics engines to construct counterfactual scenarios such as illegal cutting, complex yielding, and unconventional interactions. Meanwhile, synthetic data use should be strictly controlled, mainly using synthetic data for long-tail completion and robustness enhancement.
    \item \textbf{Optimize annotation strategies and processes.} Address multimodal spatiotemporal alignment, achieve hardware-triggered synchronization with controlled time errors, and improve geometric consistency through backend calibration optimization. Implement hierarchical annotation processes, where automatic annotation with sampling inspection to screen high-value samples is first used, followed by conducting fine annotation on selected samples, striving to improve effective annotation density and output quality without increasing overall costs.
    \item \textbf{Strengthen data version governance.} Establish version trees to manage data full lifecycle, clearly record sources, filtering rules, and applicable scope, clarifying the basis and timing for additions and removals. Map data versions one-to-one with model checkpoints, support historical effect traceability and scenario-level reproduction, improve problem localization and repair efficiency, and avoid ineffective repeated training and resource waste.
\end{itemize}

\subsection{Universal Driving Foundation Models and Regional Adaptation: Strong Transfer, Weak Customization}

E2E-AD is moving from scenario-specific optimization toward universal foundation models capable of operating across cities, road types, and driving cultures. Such models must provide consistent multimodal, long-sequence representations that align training objectives with planning and safety metrics, reducing the gap between offline accuracy and online reliability. They also need to reconcile efficient regional customization through lightweight adaptation, stable cross-domain transfer without catastrophic forgetting, and interpretable decision-making for safety assurance.

However, current foundation models degrade sharply in new domains in practice. Differences in signage, rules, and driving habits often cause planning drift, while existing adaptation methods rely on costly retraining or extensive annotation and struggle with efficiency. Performance instability and catastrophic forgetting remain common, and the lack of clear criteria for when to retrain versus fine-tune further impedes deployment. Limited interpretability compounds these issues, making adaptation outcomes difficult to assess and verify. Future research needs technological breakthroughs in the following key directions:

\begin{itemize}
    \item \textbf{Cross-domain knowledge transfer and unified representation learning.} Construct unified representation frameworks with cross-modal and cross-temporal consistency, enabling models to maintain stability across different cities and road conditions. Future work can combine contrastive learning, domain adaptation, and causal representation learning to enhance model transfer capabilities under changes in road rules, signage styles, and driving habits, thereby reducing needs for re-annotation and training.
    \item \textbf{Lightweight regional adaptation and parameter-efficient updates.} Achieve rapid updates for new cities or road styles through lightweight adaptation layers or prompt tuning methods. Research how to use small-scale incremental data for parameter low-rank updates or sparse adaptation while ensuring performance while reducing computational and storage overhead.
    \item \textbf{Stability and forgetting suppression mechanisms.} Introduce replay learning, knowledge distillation, and regularization methods in cross-domain updates to effectively suppress forgetting and reduce performance fluctuations. Meanwhile, develop dynamic judgment mechanisms and establish quantitative metrics to determine when comprehensive retraining is needed versus when only small-scale parameter adjustments are sufficient, thereby achieving efficient iteration.
\end{itemize}

\subsection{Continuous Learning and Training Operations: Closed-Loop Training Solutions with Training-Testing Integration}

To enhance E2E-AD as they move from closed sites to urban environments, it is essential to address the increasing complexity of scenarios, environmental variability, and strict safety requirements. A critical challenge arises from the persistent separation of training and testing processes: traditional methods rely on static datasets and one-time evaluations, which cannot capture the frequency and diversity of real-world changes. The resulting disconnect creates a significant gap between models that perform well in offline tests but fail to generalize effectively to unpredictable, real-world situations.

Addressing these limitations requires training-testing integration shown in Fig.~\ref{fig:training_testing_integration}.
As illustrated in the \textbf{left panel}, the process begins with \textit{data virtual-real fusion}, where real-world corner cases are augmented by controllable generation to ensure high value density.
These inputs feed into the \textbf{central Platform layer}, which manages the \textit{distillation} of large foundation models into vehicle-embedded versions.
Finally, the \textbf{right panel} depicts \textit{strategy continuous evolution}, where techniques such as evolutionary algorithms and forgetting suppression are applied to ensure that updates improve performance without degrading previously learned capabilities.
However, constructing such a system requires solving the following core tasks:

\begin{itemize}
    \item \textbf{Construct unified data-model-evaluation lineage systems.} Currently, training, testing, and deployment often belong to different platforms, lacking unified metadata structures and process records, making model performance difficult to trace and version comparisons difficult to reproduce. Future work needs to establish consistent lineage management mechanisms among training samples, model versions, and evaluation results, clarifying data inputs, optimization configurations, and validation outputs that each update depends on, supporting scenario-level reproduction verification, thereby providing clear technical basis and audit traceability for iteration paths.
    \item \textbf{Develop model evaluation systems based on closed-loop metrics.} Traditional evaluations mostly focus on accuracy metrics, making it difficult to cover strategy reasonableness and safety redundancy in end-to-end decision chains. Future work needs to construct task-level closed-loop evaluation metric systems around ``systematic stability, decision reasonableness, risk sensitivity,'' combining high-risk case replay, counterfactual perturbation verification, and simulation result comparison to achieve systematic evaluation of model behavior stability and robustness, enabling training results to truly map to online performance.
    \item \textbf{Establish reproducible minimum closed-loop mechanisms for training-validation-deployment.} Minimum training closed-loop paths centered on problem-driven, metric-constrained, and resource-sensitive approaches should be constructed. Starting from failure sample triggers, determining problem attribution through scenario replay and label verification, executing targeted retraining with minimal data increments, conducting performance confirmation through standardized validation sets, and finally supporting phased deployment and automatic rollback strategies. This mechanism needs three major characteristics: controllable processes, controllable resources, and controllable benefits, serving as underlying capabilities supporting large-scale evolution of autonomous driving.
\end{itemize}

\section{Conclusion}
\label{Conclusion}

Training constitutes the cornerstone of E2E-AD, determining both performance boundaries and the trajectory of capability evolution. While unified modeling promises advantages over modular pipelines, the training process remains constrained by long-tail data scarcity, unstable optimization, and platform bottlenecks.
The central insight of our Data-Strategy-Platform analysis is that these challenges are not isolated hurdles but systemic dependencies that require a synchronized evolution of the training ecosystem.
Future progress, therefore, depends less on individual breakthroughs in any single layer and more on the synergy across all three. It is this collaborative optimization of data quality, strategic alignment, and platform scalability that will unlock the reliability required for large-scale, sustained real-world deployment.

\ifCLASSOPTIONcaptionsoff
  \newpage
\fi

\bibliographystyle{unsrt}
\bibliography{related}

@article{chen2024end,
  title={End-to-end autonomous driving: Challenges and frontiers},
  author={Chen, Li and Wu, Penghao and Chitta, Kashyap and Jaeger, Bernhard and Geiger, Andreas and Li, Hongyang},
  journal={IEEE Transactions on Pattern Analysis and Machine Intelligence},
  year={2024},
  publisher={IEEE}
}

@article{xu2025tell,
  title={Tell-drive: Enhancing autonomous driving with teacher llm-guided deep reinforcement learning},
  author={Xu, Chengkai and Liu, Jiaqi and Fang, Shiyu and Cui, Yiming and Chen, Dong and Hang, Peng and Sun, Jian},
  journal={arXiv preprint arXiv:2502.01387},
  year={2025}
}

@article{xu2025towards,
  title={Towards Human-Centric Autonomous Driving: A Fast-Slow Architecture Integrating Large Language Model Guidance with Reinforcement Learning},
  author={Xu, Chengkai and Liu, Jiaqi and Guo, Yicheng and Zhang, Yuhang and Hang, Peng and Sun, Jian},
  journal={arXiv preprint arXiv:2505.06875},
  year={2025}
}

@article{xu2025knowledge,
  title={A Knowledge-Driven Diffusion Policy for End-to-End Autonomous Driving Based on Expert Routing},
  author={Xu, Chengkai and Liu, Jiaqi and Guo, Yicheng and Hang, Peng and Sun, Jian},
  journal={arXiv preprint arXiv:2509.04853},
  year={2025}
}

@article{feng2021intelligent,
  title={Intelligent driving intelligence test for autonomous vehicles with naturalistic and adversarial environment},
  author={Feng, Shuo and Yan, Xintao and Sun, Haowei and Feng, Yiheng and Liu, Henry X},
  journal={Nature communications},
  volume={12},
  number={1},
  pages={748},
  year={2021},
  publisher={Nature Publishing Group UK London}
}

@article{zhao2024autonomous,
  title={Autonomous driving system: A comprehensive survey},
  author={Zhao, Jingyuan and Zhao, Wenyi and Deng, Bo and Wang, Zhenghong and Zhang, Feng and Zheng, Wenxiang and Cao, Wanke and Nan, Jinrui and Lian, Yubo and Burke, Andrew F},
  journal={Expert Systems with Applications},
  volume={242},
  pages={122836},
  year={2024},
  publisher={Elsevier}
}

@article{chib2023recent,
  author={Chib, Pranav Singh and Singh, Pravendra},
  journal={IEEE Transactions on Intelligent Vehicles}, 
  title={Recent Advancements in End-to-End Autonomous Driving Using Deep Learning: A Survey}, 
  year={2024},
  volume={9},
  number={1},
  pages={103-118},
  doi={10.1109/TIV.2023.3318070}}

@article{jia2024bench2drive,
  title={Bench2drive: Towards multi-ability benchmarking of closed-loop end-to-end autonomous driving},
  author={Jia, Xiaosong and Yang, Zhenjie and Li, Qifeng and Zhang, Zhiyuan and Yan, Junchi},
  journal={Advances in Neural Information Processing Systems},
  volume={37},
  pages={819--844},
  year={2024}
}

@article{le2022survey,
  title={A survey on imitation learning techniques for end-to-end autonomous vehicles},
  author={Le Mero, Luc and Yi, Dewei and Dianati, Mehrdad and Mouzakitis, Alexandros},
  journal={IEEE Transactions on Intelligent Transportation Systems},
  volume={23},
  number={9},
  pages={14128--14147},
  year={2022},
  publisher={IEEE}
}

@article{song2023synthetic,
  title={Synthetic datasets for autonomous driving: A survey},
  author={Song, Zhihang and He, Zimin and Li, Xingyu and Ma, Qiming and Ming, Ruibo and Mao, Zhiqi and Pei, Huaxin and Peng, Lihui and Hu, Jianming and Yao, Danya and others},
  journal={IEEE Transactions on Intelligent Vehicles},
  volume={9},
  number={1},
  pages={1847--1864},
  year={2023},
  publisher={IEEE}
}

@article{liu2024survey,
  title={A survey on autonomous driving datasets: Statistics, annotation quality, and a future outlook},
  author={Liu, Mingyu and Yurtsever, Ekim and Fossaert, Jonathan and Zhou, Xingcheng and Zimmer, Walter and Cui, Yuning and Zagar, Bare Luka and Knoll, Alois C},
  journal={IEEE Transactions on Intelligent Vehicles},
  year={2024},
  publisher={IEEE}
}

@article{wang2025generative,
  title={Generative ai for autonomous driving: Frontiers and opportunities},
  author={Wang, Yuping and Xing, Shuo and Can, Cui and Li, Renjie and Hua, Hongyuan and Tian, Kexin and Mo, Zhaobin and Gao, Xiangbo and Wu, Keshu and Zhou, Sulong and others},
  journal={arXiv preprint arXiv:2505.08854},
  year={2025}
}

@inproceedings{hu2023planning,
  title={Planning-oriented autonomous driving},
  author={Hu, Yihan and Yang, Jiazhi and Chen, Li and Li, Keyu and Sima, Chonghao and Zhu, Xizhou and Chai, Siqi and Du, Senyao and Lin, Tianwei and Wang, Wenhai and others},
  booktitle={Proceedings of the IEEE/CVF conference on computer vision and pattern recognition},
  pages={17853--17862},
  year={2023}
}

@article{yan2023learning,
  title={Learning naturalistic driving environment with statistical realism},
  author={Yan, Xintao and Zou, Zhengxia and Feng, Shuo and Zhu, Haojie and Sun, Haowei and Liu, Henry X},
  journal={Nature communications},
  volume={14},
  number={1},
  pages={2037},
  year={2023},
  publisher={Nature Publishing Group UK London}
}

@inproceedings{cui2024survey,
  title={A survey on multimodal large language models for autonomous driving},
  author={Cui, Can and Ma, Yunsheng and Cao, Xu and Ye, Wenqian and Zhou, Yang and Liang, Kaizhao and Chen, Jintai and Lu, Juanwu and Yang, Zichong and Liao, Kuei-Da and others},
  booktitle={Proceedings of the IEEE/CVF winter conference on applications of computer vision},
  pages={958--979},
  year={2024}
}

@article{wang2024survey,
  title={A survey on datasets for the decision making of autonomous vehicles},
  author={Wang, Yuning and Han, Zeyu and Xing, Yining and Xu, Shaobing and Wang, Jianqiang},
  journal={IEEE Intelligent Transportation Systems Magazine},
  volume={16},
  number={2},
  pages={23--40},
  year={2024},
  publisher={IEEE}
}

@article{karangwa2023vehicle,
  title={Vehicle detection for autonomous driving: A review of algorithms and datasets},
  author={Karangwa, Jules and Liu, Jun and Zeng, Zixuan},
  journal={IEEE Transactions on Intelligent Transportation Systems},
  volume={24},
  number={11},
  pages={11568--11594},
  year={2023},
  publisher={IEEE}
}

@article{zhang2023perception,
  title={Perception and sensing for autonomous vehicles under adverse weather conditions: A survey},
  author={Zhang, Yuxiao and Carballo, Alexander and Yang, Hanting and Takeda, Kazuya},
  journal={ISPRS Journal of Photogrammetry and Remote Sensing},
  volume={196},
  pages={146--177},
  year={2023},
  publisher={Elsevier}
}

@article{xiang2023multi,
  title={Multi-sensor fusion and cooperative perception for autonomous driving: A review},
  author={Xiang, Chao and Feng, Chen and Xie, Xiaopo and Shi, Botian and Lu, Hao and Lv, Yisheng and Yang, Mingchuan and Niu, Zhendong},
  journal={IEEE Intelligent Transportation Systems Magazine},
  volume={15},
  number={5},
  pages={36--58},
  year={2023},
  publisher={IEEE}
}

@article{pomerleau1988alvinn,
  title={Alvinn: An autonomous land vehicle in a neural network},
  author={Pomerleau, Dean A},
  journal={Advances in neural information processing systems},
  volume={1},
  year={1988}
}

@article{bojarski2016end,
  title={End to end learning for self-driving cars},
  author={Bojarski, Mariusz and Del Testa, Davide and Dworakowski, Daniel and Firner, Bernhard and Flepp, Beat and Goyal, Prasoon and Jackel, Lawrence D and Monfort, Mathew and Muller, Urs and Zhang, Jiakai and others},
  journal={arXiv preprint arXiv:1604.07316},
  year={2016}
}

@inproceedings{casas2021mp3,
  title={Mp3: A unified model to map, perceive, predict and plan},
  author={Casas, Sergio and Sadat, Abbas and Urtasun, Raquel},
  booktitle={Proceedings of the IEEE/CVF Conference on Computer Vision and Pattern Recognition},
  pages={14403--14412},
  year={2021}
}

@inproceedings{karkus2023diffstack,
  title={Diffstack: A differentiable and modular control stack for autonomous vehicles},
  author={Karkus, Peter and Ivanovic, Boris and Mannor, Shie and Pavone, Marco},
  booktitle={Conference on robot learning},
  pages={2170--2180},
  year={2023},
  organization={PMLR}
}

@article{zhou2024vision,
  title={Vision language models in autonomous driving: A survey and outlook},
  author={Zhou, Xingcheng and Liu, Mingyu and Yurtsever, Ekim and Zagar, Bare Luka and Zimmer, Walter and Cao, Hu and Knoll, Alois C},
  journal={IEEE Transactions on Intelligent Vehicles},
  year={2024},
  publisher={IEEE}
}

@article{gao2024survey,
  title={A survey for foundation models in autonomous driving},
  author={Gao, Haoxiang and Wang, Zhongruo and Li, Yaqian and Long, Kaiwen and Yang, Ming and Shen, Yiqing},
  journal={arXiv preprint arXiv:2402.01105},
  year={2024}
}

@inproceedings{shao2024lmdrive,
  title={Lmdrive: Closed-loop end-to-end driving with large language models},
  author={Shao, Hao and Hu, Yuxuan and Wang, Letian and Song, Guanglu and Waslander, Steven L and Liu, Yu and Li, Hongsheng},
  booktitle={Proceedings of the IEEE/CVF Conference on Computer Vision and Pattern Recognition},
  pages={15120--15130},
  year={2024}
}

@inproceedings{sima2024drivelm,
  title={Drivelm: Driving with graph visual question answering},
  author={Sima, Chonghao and Renz, Katrin and Chitta, Kashyap and Chen, Li and Zhang, Hanxue and Xie, Chengen and Bei{\ss}wenger, Jens and Luo, Ping and Geiger, Andreas and Li, Hongyang},
  booktitle={European conference on computer vision},
  pages={256--274},
  year={2024},
  organization={Springer}
}

@article{zhou2025opendrivevla,
  title={Opendrivevla: Towards end-to-end autonomous driving with large vision language action model},
  author={Zhou, Xingcheng and Han, Xuyuan and Yang, Feng and Ma, Yunpu and Knoll, Alois C},
  journal={arXiv preprint arXiv:2503.23463},
  year={2025}
}

@inproceedings{wang2024drivedreamer,
  title={Drivedreamer: Towards real-world-drive world models for autonomous driving},
  author={Wang, Xiaofeng and Zhu, Zheng and Huang, Guan and Chen, Xinze and Zhu, Jiagang and Lu, Jiwen},
  booktitle={European conference on computer vision},
  pages={55--72},
  year={2024},
  organization={Springer}
}

@inproceedings{zhao2025drivedreamer,
  title={Drivedreamer-2: Llm-enhanced world models for diverse driving video generation},
  author={Zhao, Guosheng and Wang, Xiaofeng and Zhu, Zheng and Chen, Xinze and Huang, Guan and Bao, Xiaoyi and Wang, Xingang},
  booktitle={Proceedings of the AAAI Conference on Artificial Intelligence},
  volume={39},
  number={10},
  pages={10412--10420},
  year={2025}
}

@article{hu2023gaia,
  title={Gaia-1: A generative world model for autonomous driving},
  author={Hu, Anthony and Russell, Lloyd and Yeo, Hudson and Murez, Zak and Fedoseev, George and Kendall, Alex and Shotton, Jamie and Corrado, Gianluca},
  journal={arXiv preprint arXiv:2309.17080},
  year={2023}
}

@article{russell2025gaia,
  title={Gaia-2: A controllable multi-view generative world model for autonomous driving},
  author={Russell, Lloyd and Hu, Anthony and Bertoni, Lorenzo and Fedoseev, George and Shotton, Jamie and Arani, Elahe and Corrado, Gianluca},
  journal={arXiv preprint arXiv:2503.20523},
  year={2025}
}

@article{tian2024tokenize,
  title={Tokenize the world into object-level knowledge to address long-tail events in autonomous driving},
  author={Tian, Ran and Li, Boyi and Weng, Xinshuo and Chen, Yuxiao and Schmerling, Edward and Wang, Yue and Ivanovic, Boris and Pavone, Marco},
  journal={arXiv preprint arXiv:2407.00959},
  year={2024}
}

@article{jiang2025naturalistic,
  title={A naturalistic trajectory dataset with dense interaction for autonomous driving},
  author={Jiang, Xiyan and Zhao, Xiaocong and Liu, Yiru and Li, Zirui and Hang, Peng and Xiong, Lu and Sun, Jian},
  journal={Scientific Data},
  volume={12},
  number={1},
  pages={1084},
  year={2025},
  publisher={Nature Publishing Group UK London}
}

@inproceedings{yang2024human,
  title={Human-centric autonomous systems with llms for user command reasoning},
  author={Yang, Yi and Zhang, Qingwen and Li, Ci and Marta, Daniel Sim{\~o}es and Batool, Nazre and Folkesson, John},
  booktitle={Proceedings of the IEEE/CVF Winter Conference on Applications of Computer Vision},
  pages={988--994},
  year={2024}
}

@article{schrum2024maveric,
  title={Maveric: A data-driven approach to personalized autonomous driving},
  author={Schrum, Mariah L and Sumner, Emily and Gombolay, Matthew C and Best, Andrew},
  journal={IEEE Transactions on Robotics},
  volume={40},
  pages={1952--1965},
  year={2024},
  publisher={IEEE}
}

@article{caesar2021nuplan,
  title={nuplan: A closed-loop ml-based planning benchmark for autonomous vehicles},
  author={Caesar, Holger and Kabzan, Juraj and Tan, Kok Seang and Fong, Whye Kit and Wolff, Eric and Lang, Alex and Fletcher, Luke and Beijbom, Oscar and Omari, Sammy},
  journal={arXiv preprint arXiv:2106.11810},
  year={2021}
}

@inproceedings{zhang2022rethinking,
  title={Rethinking closed-loop training for autonomous driving},
  author={Zhang, Chris and Guo, Runsheng and Zeng, Wenyuan and Xiong, Yuwen and Dai, Binbin and Hu, Rui and Ren, Mengye and Urtasun, Raquel},
  booktitle={European Conference on Computer Vision},
  pages={264--282},
  year={2022},
  organization={Springer}
}

@inproceedings{dosovitskiy2017carla,
  title={CARLA: An open urban driving simulator},
  author={Dosovitskiy, Alexey and Ros, German and Codevilla, Felipe and Lopez, Antonio and Koltun, Vladlen},
  booktitle={Conference on robot learning},
  pages={1--16},
  year={2017},
  organization={PMLR}
}

@article{jacobs2023deepspeed,
  title={Deepspeed ulysses: System optimizations for enabling training of extreme long sequence transformer models},
  author={Jacobs, Sam Ade and Tanaka, Masahiro and Zhang, Chengming and Zhang, Minjia and Song, Shuaiwen Leon and Rajbhandari, Samyam and He, Yuxiong},
  journal={arXiv preprint arXiv:2309.14509},
  year={2023}
}

@article{miao2023towards,
  title={Towards efficient generative large language model serving: A survey from algorithms to systems},
  author={Miao, Xupeng and Oliaro, Gabriele and Zhang, Zhihao and Cheng, Xinhao and Jin, Hongyi and Chen, Tianqi and Jia, Zhihao},
  journal={ACM Computing Surveys},
  year={2023},
  publisher={ACM New York, NY}
}

@article{zheng2024llamafactory,
  title={Llamafactory: Unified efficient fine-tuning of 100+ language models},
  author={Zheng, Yaowei and Zhang, Richong and Zhang, Junhao and Ye, Yanhan and Luo, Zheyan and Feng, Zhangchi and Ma, Yongqiang},
  journal={arXiv preprint arXiv:2403.13372},
  year={2024}
}

@article{zhang2024found,
  title={Found in the middle: How language models use long contexts better via plug-and-play positional encoding},
  author={Zhang, Zhenyu and Chen, Runjin and Liu, Shiwei and Yao, Zhewei and Ruwase, Olatunji and Chen, Beidi and Wu, Xiaoxia and Wang, Zhangyang and others},
  journal={Advances in Neural Information Processing Systems},
  volume={37},
  pages={60755--60775},
  year={2024}
}

@article{cai2025survey,
  title={A survey on mixture of experts in large language models},
  author={Cai, Weilin and Jiang, Juyong and Wang, Fan and Tang, Jing and Kim, Sunghun and Huang, Jiayi},
  journal={IEEE Transactions on Knowledge and Data Engineering},
  year={2025},
  publisher={IEEE}
}

@article{lepikhin2020gshard,
  title={Gshard: Scaling giant models with conditional computation and automatic sharding},
  author={Lepikhin, Dmitry and Lee, HyoukJoong and Xu, Yuanzhong and Chen, Dehao and Firat, Orhan and Huang, Yanping and Krikun, Maxim and Shazeer, Noam and Chen, Zhifeng},
  journal={arXiv preprint arXiv:2006.16668},
  year={2020}
}

@inproceedings{rajbhandari2022deepspeed,
  title={Deepspeed-moe: Advancing mixture-of-experts inference and training to power next-generation ai scale},
  author={Rajbhandari, Samyam and Li, Conglong and Yao, Zhewei and Zhang, Minjia and Aminabadi, Reza Yazdani and Awan, Ammar Ahmad and Rasley, Jeff and He, Yuxiong},
  booktitle={International conference on machine learning},
  pages={18332--18346},
  year={2022},
  organization={PMLR}
}

@inproceedings{rajbhandari2020zero,
  title={Zero: Memory optimizations toward training trillion parameter models},
  author={Rajbhandari, Samyam and Rasley, Jeff and Ruwase, Olatunji and He, Yuxiong},
  booktitle={SC20: International Conference for High Performance Computing, Networking, Storage and Analysis},
  pages={1--16},
  year={2020},
  organization={IEEE}
}

@article{sergeev2018horovod,
  title={Horovod: fast and easy distributed deep learning in TensorFlow},
  author={Sergeev, Alexander and Del Balso, Mike},
  journal={arXiv preprint arXiv:1802.05799},
  year={2018}
}

@inproceedings{subhlok1993exploiting,
  title={Exploiting task and data parallelism on a multicomputer},
  author={Subhlok, Jaspal and Stichnoth, James M and O'hallaron, David R and Gross, Thomas},
  booktitle={Proceedings of the fourth ACM SIGPLAN symposium on Principles and practice of parallel programming},
  pages={13--22},
  year={1993}
}

@article{cortes2021object,
  title={Object detection, distributed cloud computing and parallelization techniques for autonomous driving systems},
  author={Cortes Gallardo Medina, Edgar and Velazquez Espitia, Victor Miguel and Chipuli Silva, Daniela and Fernandez Ruiz de las Cuevas, Sebastian and Palacios Hirata, Marco and Zhu Chen, Alfredo and Gonzalez Gonzalez, Jose Angel and Bustamante-Bello, Rogelio and Moreno-Garcia, Carlos Francisco},
  journal={Applied sciences},
  volume={11},
  number={7},
  pages={2925},
  year={2021},
  publisher={MDPI}
}

@inproceedings{narayanan2021memory,
  title={Memory-efficient pipeline-parallel dnn training},
  author={Narayanan, Deepak and Phanishayee, Amar and Shi, Kaiyu and Chen, Xie and Zaharia, Matei},
  booktitle={International Conference on Machine Learning},
  pages={7937--7947},
  year={2021},
  organization={PMLR}
}

@inproceedings{narayanan2021efficient,
  title={Efficient large-scale language model training on gpu clusters using megatron-lm},
  author={Narayanan, Deepak and Shoeybi, Mohammad and Casper, Jared and LeGresley, Patrick and Patwary, Mostofa and Korthikanti, Vijay and Vainbrand, Dmitri and Kashinkunti, Prethvi and Bernauer, Julie and Catanzaro, Bryan and others},
  booktitle={Proceedings of the international conference for high performance computing, networking, storage and analysis},
  pages={1--15},
  year={2021}
}

@inproceedings{wang2022tesseract,
  title={Tesseract: Parallelize the tensor parallelism efficiently},
  author={Wang, Boxiang and Xu, Qifan and Bian, Zhengda and You, Yang},
  booktitle={Proceedings of the 51st International Conference on Parallel Processing},
  pages={1--11},
  year={2022}
}

@article{li2021sequence,
  title={Sequence parallelism: Long sequence training from system perspective},
  author={Li, Shenggui and Xue, Fuzhao and Baranwal, Chaitanya and Li, Yongbin and You, Yang},
  journal={arXiv preprint arXiv:2105.13120},
  year={2021}
}

@inproceedings{sun2024adapipe,
  title={Adapipe: Optimizing pipeline parallelism with adaptive recomputation and partitioning},
  author={Sun, Zhenbo and Cao, Huanqi and Wang, Yuanwei and Feng, Guanyu and Chen, Shengqi and Wang, Haojie and Chen, Wenguang},
  booktitle={Proceedings of the 29th ACM International Conference on Architectural Support for Programming Languages and Operating Systems, Volume 3},
  pages={86--100},
  year={2024}
}

@inproceedings{kim2023bpipe,
  title={Bpipe: Memory-balanced pipeline parallelism for training large language models},
  author={Kim, Taebum and Kim, Hyoungjoo and Yu, Gyeong-In and Chun, Byung-Gon},
  booktitle={International Conference on Machine Learning},
  pages={16639--16653},
  year={2023},
  organization={PMLR}
}

@inproceedings{liu2023hanayo,
  title={Hanayo: Harnessing wave-like pipeline parallelism for enhanced large model training efficiency},
  author={Liu, Ziming and Cheng, Shenggan and Zhou, Haotian and You, Yang},
  booktitle={Proceedings of the International Conference for High Performance Computing, Networking, Storage and Analysis},
  pages={1--13},
  year={2023}
}

@article{verwimp2023clad,
  title={Clad: A realistic continual learning benchmark for autonomous driving},
  author={Verwimp, Eli and Yang, Kuo and Parisot, Sarah and Hong, Lanqing and McDonagh, Steven and P{\'e}rez-Pellitero, Eduardo and De Lange, Matthias and Tuytelaars, Tinne},
  journal={Neural Networks},
  volume={161},
  pages={659--669},
  year={2023},
  publisher={Elsevier}
}

@article{li2024platform,
  title={A platform-agnostic deep reinforcement learning framework for effective Sim2Real transfer towards autonomous driving},
  author={Li, Dianzhao and Okhrin, Ostap},
  journal={Communications Engineering},
  volume={3},
  number={1},
  pages={147},
  year={2024},
  publisher={Nature Publishing Group UK London}
}

@article{li2024data,
  title={Data-centric evolution in autonomous driving: A comprehensive survey of big data system, data mining, and closed-loop technologies},
  author={Li, Lincan and Shao, Wei and Dong, Wei and Tian, Yijun and Zhang, Qiming and Yang, Kaixiang and Zhang, Wenjie},
  journal={arXiv preprint arXiv:2401.12888},
  year={2024}
}

@article{ge2024life,
  title={Life-long Learning and Testing for Automated Vehicles via Adaptive Scenario Sampling as A Continuous Optimization Process},
  author={Ge, Jingwei and Wang, Pengbo and Chang, Cheng and Zhang, Yi and Yao, Danya and Li, Li},
  journal={IEEE Transactions on Intelligent Vehicles},
  year={2024},
  publisher={IEEE}
}

@inproceedings{geiger2012we,
  title={Are we ready for autonomous driving? the kitti vision benchmark suite},
  author={Geiger, Andreas and Lenz, Philip and Urtasun, Raquel},
  booktitle={2012 IEEE conference on computer vision and pattern recognition},
  pages={3354--3361},
  year={2012},
  organization={IEEE}
}

@article{wilson2023argoverse,
  title={Argoverse 2: Next generation datasets for self-driving perception and forecasting},
  author={Wilson, Benjamin and Qi, William and Agarwal, Tanmay and Lambert, John and Singh, Jagjeet and Khandelwal, Siddhesh and Pan, Bowen and Kumar, Ratnesh and Hartnett, Andrew and Pontes, Jhony Kaesemodel and others},
  journal={arXiv preprint arXiv:2301.00493},
  year={2023}
}

@article{mao2021one,
  title={One million scenes for autonomous driving: Once dataset},
  author={Mao, Jiageng and Niu, Minzhe and Jiang, Chenhan and Liang, Hanxue and Chen, Jingheng and Liang, Xiaodan and Li, Yamin and Ye, Chaoqiang and Zhang, Wei and Li, Zhenguo and others},
  journal={arXiv preprint arXiv:2106.11037},
  year={2021}
}

@inproceedings{sun2020scalability,
  title={Scalability in perception for autonomous driving: Waymo open dataset},
  author={Sun, Pei and Kretzschmar, Henrik and Dotiwalla, Xerxes and Chouard, Aurelien and Patnaik, Vijaysai and Tsui, Paul and Guo, James and Zhou, Yin and Chai, Yuning and Caine, Benjamin and others},
  booktitle={Proceedings of the IEEE/CVF conference on computer vision and pattern recognition},
  pages={2446--2454},
  year={2020}
}

@inproceedings{alibeigi2023zenseact,
  title={Zenseact open dataset: A large-scale and diverse multimodal dataset for autonomous driving},
  author={Alibeigi, Mina and Ljungbergh, William and Tonderski, Adam and Hess, Georg and Lilja, Adam and Lindstr{\"o}m, Carl and Motorniuk, Daria and Fu, Junsheng and Widahl, Jenny and Petersson, Christoffer},
  booktitle={Proceedings of the IEEE/CVF International Conference on Computer Vision},
  pages={20178--20188},
  year={2023}
}

@inproceedings{hendrycks2022scaling,
  title={Scaling Out-of-Distribution Detection for Real-World Settings},
  author={Hendrycks, Dan and Basart, Steven and Mazeika, Mantas and Zou, Andy and Kwon, Joseph and Mostajabi, Mohammadreza and Steinhardt, Jacob and Song, Dawn},
  booktitle={International Conference on Machine Learning},
  pages={8759--8773},
  year={2022},
  organization={PMLR}
}

@inproceedings{sun2022shift,
  title={SHIFT: a synthetic driving dataset for continuous multi-task domain adaptation},
  author={Sun, Tao and Segu, Mattia and Postels, Janis and Wang, Yuxuan and Van Gool, Luc and Schiele, Bernt and Tombari, Federico and Yu, Fisher},
  booktitle={Proceedings of the IEEE/CVF conference on computer vision and pattern recognition},
  pages={21371--21382},
  year={2022}
}

@inproceedings{xu2022v2x,
  title={V2x-vit: Vehicle-to-everything cooperative perception with vision transformer},
  author={Xu, Runsheng and Xiang, Hao and Tu, Zhengzhong and Xia, Xin and Yang, Ming-Hsuan and Ma, Jiaqi},
  booktitle={European conference on computer vision},
  pages={107--124},
  year={2022},
  organization={Springer}
}

@inproceedings{wang2024deepaccident,
  title={Deepaccident: A motion and accident prediction benchmark for v2x autonomous driving},
  author={Wang, Tianqi and Kim, Sukmin and Wenxuan, Ji and Xie, Enze and Ge, Chongjian and Chen, Junsong and Li, Zhenguo and Luo, Ping},
  booktitle={Proceedings of the AAAI Conference on Artificial Intelligence},
  volume={38},
  number={6},
  pages={5599--5606},
  year={2024}
}

@inproceedings{huang2018apolloscape,
  title={The apolloscape dataset for autonomous driving},
  author={Huang, Xinyu and Cheng, Xinjing and Geng, Qichuan and Cao, Binbin and Zhou, Dingfu and Wang, Peng and Lin, Yuanqing and Yang, Ruigang},
  booktitle={Proceedings of the IEEE conference on computer vision and pattern recognition workshops},
  pages={954--960},
  year={2018}
}

@article{longo2017crowd,
  title={Crowd-sourced data collection for urban monitoring via mobile sensors},
  author={Longo, Antonella and Zappatore, Marco and Bochicchio, Mario and Navathe, Shamkant B},
  journal={ACM Transactions on Internet Technology (TOIT)},
  volume={18},
  number={1},
  pages={1--21},
  year={2017},
  publisher={ACM New York, NY, USA}
}

@article{xia2023automated,
  title={An automated driving systems data acquisition and analytics platform},
  author={Xia, Xin and Meng, Zonglin and Han, Xu and Li, Hanzhao and Tsukiji, Takahiro and Xu, Runsheng and Zheng, Zhaoliang and Ma, Jiaqi},
  journal={Transportation research part C: emerging technologies},
  volume={151},
  pages={104120},
  year={2023},
  publisher={Elsevier}
}

@inproceedings{silva2022ride,
  title={Ride-hailing service aware electric taxi fleet management using reinforcement learning},
  author={Silva, Paul and Han, Young Joo and Kim, Young-Chon and Kang, Dong-Ki},
  booktitle={2022 Thirteenth International Conference on Ubiquitous and Future Networks (ICUFN)},
  pages={427--432},
  year={2022},
  organization={IEEE}
}

@inproceedings{zimmer2024tumtraf,
  title={Tumtraf v2x cooperative perception dataset},
  author={Zimmer, Walter and Wardana, Gerhard Arya and Sritharan, Suren and Zhou, Xingcheng and Song, Rui and Knoll, Alois C},
  booktitle={Proceedings of the IEEE/CVF conference on computer vision and pattern recognition},
  pages={22668--22677},
  year={2024}
}

@inproceedings{zimmer2023tumtraf,
  title={Tumtraf intersection dataset: All you need for urban 3d camera-lidar roadside perception},
  author={Zimmer, Walter and Cre{\ss}, Christian and Nguyen, Huu Tung and Knoll, Alois C},
  booktitle={2023 IEEE 26th International Conference on Intelligent Transportation Systems (ITSC)},
  pages={1030--1037},
  year={2023},
  organization={IEEE}
}

@inproceedings{xu2023v2v4real,
  title={V2v4real: A real-world large-scale dataset for vehicle-to-vehicle cooperative perception},
  author={Xu, Runsheng and Xia, Xin and Li, Jinlong and Li, Hanzhao and Zhang, Shuo and Tu, Zhengzhong and Meng, Zonglin and Xiang, Hao and Dong, Xiaoyu and Song, Rui and others},
  booktitle={Proceedings of the IEEE/CVF conference on computer vision and pattern recognition},
  pages={13712--13722},
  year={2023}
}

@inproceedings{ye2022rope3d,
  title={Rope3d: The roadside perception dataset for autonomous driving and monocular 3d object detection task},
  author={Ye, Xiaoqing and Shu, Mao and Li, Hanyu and Shi, Yifeng and Li, Yingying and Wang, Guangjie and Tan, Xiao and Ding, Errui},
  booktitle={Proceedings of the IEEE/CVF Conference on Computer Vision and Pattern Recognition},
  pages={21341--21350},
  year={2022}
}

@inproceedings{busch2022lumpi,
  title={Lumpi: The leibniz university multi-perspective intersection dataset},
  author={Busch, Steffen and Koetsier, Christian and Axmann, Jeldrik and Brenner, Claus},
  booktitle={2022 IEEE Intelligent Vehicles Symposium (IV)},
  pages={1127--1134},
  year={2022},
  organization={IEEE}
}

@article{qiao2024trafficgamer,
  title={TrafficGamer: Reliable and Flexible Traffic Simulation for Safety-Critical Scenarios with Game-Theoretic Oracles},
  author={Qiao, Guanren and Quan, Guorui and Yu, Jiawei and Jia, Shujun and Liu, Guiliang},
  journal={arXiv preprint arXiv:2408.15538},
  year={2024}
}

@article{guo2025interactive,
  title={Interactive Adversarial Testing of Autonomous Vehicles with Adjustable Confrontation Intensity},
  author={Guo, Yicheng and Xu, Chengkai and Liu, Jiaqi and Zhang, Hao and Hang, Peng and Sun, Jian},
  journal={arXiv preprint arXiv:2507.21814},
  year={2025}
}

@article{chen2025s,
  title={S-nerf++: Autonomous driving simulation via neural reconstruction and generation},
  author={Chen, Yurui and Zhang, Junge and Xie, Ziyang and Li, Wenye and Zhang, Feihu and Lu, Jiachen and Zhang, Li},
  journal={IEEE Transactions on Pattern Analysis and Machine Intelligence},
  year={2025},
  publisher={IEEE}
}

@article{zhu2024scene,
  title={Scene reconstruction techniques for autonomous driving: a review of 3D Gaussian splatting},
  author={Zhu, Huixin and Zhang, Zhili and Zhao, Junyang and Duan, Hui and Ding, Yao and Xiao, Xiongwu and Yuan, Junsong},
  journal={Artificial Intelligence Review},
  volume={58},
  number={1},
  pages={30},
  year={2024},
  publisher={Springer}
}

@article{liao2025learning,
  title={Learning-based 3D Reconstruction in Autonomous Driving: A Comprehensive Survey},
  author={Liao, Liewen and Yan, Weihao and Yang, Ming and Zhang, Songan},
  journal={arXiv preprint arXiv:2503.14537},
  year={2025}
}

@article{guan2024world,
  title={World models for autonomous driving: An initial survey},
  author={Guan, Yanchen and Liao, Haicheng and Li, Zhenning and Hu, Jia and Yuan, Runze and Zhang, Guohui and Xu, Chengzhong},
  journal={IEEE Transactions on Intelligent Vehicles},
  year={2024},
  publisher={IEEE}
}

@inproceedings{wang2024driving,
  title={Driving into the future: Multiview visual forecasting and planning with world model for autonomous driving},
  author={Wang, Yuqi and He, Jiawei and Fan, Lue and Li, Hongxin and Chen, Yuntao and Zhang, Zhaoxiang},
  booktitle={Proceedings of the IEEE/CVF Conference on Computer Vision and Pattern Recognition},
  pages={14749--14759},
  year={2024}
}

@inproceedings{lu2024scenecontrol,
  title={Scenecontrol: Diffusion for controllable traffic scene generation},
  author={Lu, Jack and Wong, Kelvin and Zhang, Chris and Suo, Simon and Urtasun, Raquel},
  booktitle={2024 IEEE International Conference on Robotics and Automation (ICRA)},
  pages={16908--16914},
  year={2024},
  organization={IEEE}
}

@inproceedings{xu2025diffscene,
  title={Diffscene: Diffusion-based safety-critical scenario generation for autonomous vehicles},
  author={Xu, Chejian and Petiushko, Aleksandr and Zhao, Ding and Li, Bo},
  booktitle={Proceedings of the AAAI Conference on Artificial Intelligence},
  volume={39},
  number={8},
  pages={8797--8805},
  year={2025}
}

@article{bai2024bridging,
  title={Bridging the domain gap between synthetic and real-world data for autonomous driving},
  author={Bai, Xiangyu and Luo, Yedi and Jiang, Le and Gupta, Aniket and Kaveti, Pushyami and Singh, Hanumant and Ostadabbas, Sarah},
  journal={Journal on Autonomous Transportation Systems},
  volume={1},
  number={2},
  pages={1--15},
  year={2024},
  publisher={ACM New York, NY}
}

@inproceedings{cheng2022masked,
  title={Masked-attention mask transformer for universal image segmentation},
  author={Cheng, Bowen and Misra, Ishan and Schwing, Alexander G and Kirillov, Alexander and Girdhar, Rohit},
  booktitle={Proceedings of the IEEE/CVF conference on computer vision and pattern recognition},
  pages={1290--1299},
  year={2022}
}

@article{cheng2021per,
  title={Per-pixel classification is not all you need for semantic segmentation},
  author={Cheng, Bowen and Schwing, Alex and Kirillov, Alexander},
  journal={Advances in neural information processing systems},
  volume={34},
  pages={17864--17875},
  year={2021}
}

@article{hirose2025learning,
  title={Learning to Drive Anywhere with Model-Based Reannotation},
  author={Hirose, Noriaki and Ignatova, Lydia and Stachowicz, Kyle and Glossop, Catherine and Levine, Sergey and Shah, Dhruv},
  journal={arXiv preprint arXiv:2505.05592},
  year={2025}
}

@article{gui2024survey,
  title={A survey on self-supervised learning: Algorithms, applications, and future trends},
  author={Gui, Jie and Chen, Tuo and Zhang, Jing and Cao, Qiong and Sun, Zhenan and Luo, Hao and Tao, Dacheng},
  journal={IEEE Transactions on Pattern Analysis and Machine Intelligence},
  volume={46},
  number={12},
  pages={9052--9071},
  year={2024},
  publisher={IEEE}
}

@inproceedings{li2024drivingdiffusion,
  title={DrivingDiffusion: layout-guided multi-view driving scenarios video generation with latent diffusion model},
  author={Li, Xiaofan and Zhang, Yifu and Ye, Xiaoqing},
  booktitle={European Conference on Computer Vision},
  pages={469--485},
  year={2024},
  organization={Springer}
}

@inproceedings{song2022scenario,
  title={A scenario distribution model for effective and efficient testing of autonomous driving systems},
  author={Song, Qunying and Runeson, Per and Persson, Stefan},
  booktitle={Proceedings of the 37th IEEE/ACM International Conference on Automated Software Engineering},
  pages={1--8},
  year={2022}
}

@article{xie2024diversified,
  title={Diversified Critical-Scenario-Search for Defect Detection of Autonomous Driving System},
  author={Xie, Yizhou and Zhang, Yong and Yin, Chengliang and Huang, Hongcheng and Dai, Kunpeng},
  journal={IEEE Transactions on Intelligent Vehicles},
  year={2024},
  publisher={IEEE}
}

@article{yang2025quantitative,
  title={Quantitative representation of autonomous driving scenario difficulty based on adversarial policy search},
  author={Yang, Shuo and Wang, Caojun and Zhang, Yuanjian and Yin, Yuming and Huang, Yanjun and Li, Shengbo Eben and Chen, Hong},
  journal={Research},
  volume={8},
  pages={0575},
  year={2025},
  publisher={AAAS}
}

@article{sheng2025curricuvlm,
  title={Curricuvlm: Towards safe autonomous driving via personalized safety-critical curriculum learning with vision-language models},
  author={Sheng, Zihao and Huang, Zilin and Qu, Yansong and Leng, Yue and Bhavanam, Sruthi and Chen, Sikai},
  journal={arXiv preprint arXiv:2502.15119},
  year={2025}
}

@article{gheisari2024cappad,
  title={CAPPAD: a privacy-preservation solution for autonomous vehicles using SDN, differential privacy and data aggregation},
  author={Gheisari, Mehdi and Khan, Wazir Zada and Najafabadi, Hamid Esmaeili and McArdle, Gavin and Rabiei-Dastjerdi, Hamidreza and Liu, Yang and Fern{\'a}ndez-Campusano, Christian and Abdalla, Hemn Barzan},
  journal={Applied Intelligence},
  volume={54},
  number={4},
  pages={3417--3428},
  year={2024},
  publisher={Springer Nature BV}
}

@article{wu2023add,
  title={ADD: An automatic desensitization fisheye dataset for autonomous driving},
  author={Wu, Zizhang and Chen, Xinyuan and Wei, Hongyang and Song, Fan and Xu, Tianhao},
  journal={Engineering Applications of Artificial Intelligence},
  volume={126},
  pages={106766},
  year={2023},
  publisher={Elsevier}
}

@article{hussein2017imitation,
  title={Imitation learning: A survey of learning methods},
  author={Hussein, Ahmed and Gaber, Mohamed Medhat and Elyan, Eyad and Jayne, Chrisina},
  journal={ACM Computing Surveys (CSUR)},
  volume={50},
  number={2},
  pages={1--35},
  year={2017},
  publisher={ACM New York, NY, USA}
}

@article{ho2016generative,
  title={Generative adversarial imitation learning},
  author={Ho, Jonathan and Ermon, Stefano},
  journal={Advances in neural information processing systems},
  volume={29},
  year={2016}
}

@inproceedings{hawke2020urban,
  title={Urban driving with conditional imitation learning},
  author={Hawke, Jeffrey and Shen, Richard and Gurau, Corina and Sharma, Siddharth and Reda, Daniele and Nikolov, Nikolay and Mazur, Przemys{\l}aw and Micklethwaite, Sean and Griffiths, Nicolas and Shah, Amar and others},
  booktitle={2020 IEEE International Conference on Robotics and Automation (ICRA)},
  pages={251--257},
  year={2020},
  organization={IEEE}
}

@inproceedings{prakash2020exploring,
  title={Exploring data aggregation in policy learning for vision-based urban autonomous driving},
  author={Prakash, Aditya and Behl, Aseem and Ohn-Bar, Eshed and Chitta, Kashyap and Geiger, Andreas},
  booktitle={Proceedings of the IEEE/CVF Conference on Computer Vision and Pattern Recognition},
  pages={11763--11773},
  year={2020}
}

@article{diehl2023uncertainty,
  title={Uncertainty-aware model-based offline reinforcement learning for automated driving},
  author={Diehl, Christopher and Sievernich, Timo Sebastian and Kr{\"u}ger, Martin and Hoffmann, Frank and Bertram, Torsten},
  journal={IEEE Robotics and Automation Letters},
  volume={8},
  number={2},
  pages={1167--1174},
  year={2023},
  publisher={IEEE}
}

@inproceedings{caesar2020nuscenes,
  title={nuscenes: A multimodal dataset for autonomous driving},
  author={Caesar, Holger and Bankiti, Varun and Lang, Alex H and Vora, Sourabh and Liong, Venice Erin and Xu, Qiang and Krishnan, Anush and Pan, Yu and Baldan, Giancarlo and Beijbom, Oscar},
  booktitle={Proceedings of the IEEE/CVF conference on computer vision and pattern recognition},
  pages={11621--11631},
  year={2020}
}

@inproceedings{lee2024ad4rl,
  title={AD4RL: Autonomous driving benchmarks for offline reinforcement learning with value-based dataset},
  author={Lee, Dongsu and Eom, Chanin and Kwon, Minhae},
  booktitle={2024 IEEE International Conference on Robotics and Automation (ICRA)},
  pages={8239--8245},
  year={2024},
  organization={IEEE}
}

@article{lin2024safety,
  title={Safety-aware causal representation for trustworthy offline reinforcement learning in autonomous driving},
  author={Lin, Haohong and Ding, Wenhao and Liu, Zuxin and Niu, Yaru and Zhu, Jiacheng and Niu, Yuming and Zhao, Ding},
  journal={IEEE Robotics and Automation Letters},
  volume={9},
  number={5},
  pages={4639--4646},
  year={2024},
  publisher={IEEE}
}

@inproceedings{li2024boosting,
  title={Boosting offline reinforcement learning for autonomous driving with hierarchical latent skills},
  author={Li, Zenan and Nie, Fan and Sun, Qiao and Da, Fang and Zhao, Hang},
  booktitle={2024 IEEE International Conference on Robotics and Automation (ICRA)},
  pages={18362--18369},
  year={2024},
  organization={IEEE}
}

@inproceedings{lu2023imitation,
  title={Imitation is not enough: Robustifying imitation with reinforcement learning for challenging driving scenarios},
  author={Lu, Yiren and Fu, Justin and Tucker, George and Pan, Xinlei and Bronstein, Eli and Roelofs, Rebecca and Sapp, Benjamin and White, Brandyn and Faust, Aleksandra and Whiteson, Shimon and others},
  booktitle={2023 IEEE/RSJ International Conference on Intelligent Robots and Systems (IROS)},
  pages={7553--7560},
  year={2023},
  organization={IEEE}
}

@inproceedings{zhu2024decoupled,
  title={Decoupled deep reinforcement learning with sensor fusion and imitation learning for autonomous driving optimization},
  author={Zhu, Junlin and Ortiz, Jorge and Sun, Yuan},
  booktitle={2024 6th International Conference on Artificial Intelligence and Computer Applications (ICAICA)},
  pages={306--310},
  year={2024},
  organization={IEEE}
}

@inproceedings{duan2024prompting,
  title={Prompting multi-modal tokens to enhance end-to-end autonomous driving imitation learning with llms},
  author={Duan, Yiqun and Zhang, Qiang and Xu, Renjing},
  booktitle={2024 IEEE International Conference on Robotics and Automation (ICRA)},
  pages={6798--6805},
  year={2024},
  organization={IEEE}
}

@inproceedings{yan2025streetcrafter,
  title={Streetcrafter: Street view synthesis with controllable video diffusion models},
  author={Yan, Yunzhi and Xu, Zhen and Lin, Haotong and Jin, Haian and Guo, Haoyu and Wang, Yida and Zhan, Kun and Lang, Xianpeng and Bao, Hujun and Zhou, Xiaowei and others},
  booktitle={Proceedings of the Computer Vision and Pattern Recognition Conference},
  pages={822--832},
  year={2025}
}

@inproceedings{yan2025drivingsphere,
  title={Drivingsphere: Building a high-fidelity 4d world for closed-loop simulation},
  author={Yan, Tianyi and Wu, Dongming and Han, Wencheng and Jiang, Junpeng and Zhou, Xia and Zhan, Kun and Xu, Cheng-zhong and Shen, Jianbing},
  booktitle={Proceedings of the Computer Vision and Pattern Recognition Conference},
  pages={27531--27541},
  year={2025}
}

@inproceedings{ni2025recondreamer,
  title={Recondreamer: Crafting world models for driving scene reconstruction via online restoration},
  author={Ni, Chaojun and Zhao, Guosheng and Wang, Xiaofeng and Zhu, Zheng and Qin, Wenkang and Huang, Guan and Liu, Chen and Chen, Yuyin and Wang, Yida and Zhang, Xueyang and others},
  booktitle={Proceedings of the Computer Vision and Pattern Recognition Conference},
  pages={1559--1569},
  year={2025}
}

@article{gu2023safe,
  title={Safe-state enhancement method for autonomous driving via direct hierarchical reinforcement learning},
  author={Gu, Ziqing and Gao, Lingping and Ma, Haitong and Li, Shengbo Eben and Zheng, Sifa and Jing, Wei and Chen, Junbo},
  journal={IEEE Transactions on Intelligent Transportation Systems},
  volume={24},
  number={9},
  pages={9966--9983},
  year={2023},
  publisher={IEEE}
}

@article{hou2024safe,
  title={Safe reinforcement learning for autonomous driving by using disturbance-observer-based control barrier functions},
  author={Hou, Zhengyu and Liu, Wenjun and Knoll, Alois},
  journal={IEEE Transactions on Intelligent Vehicles},
  year={2024},
  publisher={IEEE}
}

@article{zhang2025control,
  title={Control Barrier Function-Guided Deep Reinforcement Learning for Decision-Making of Autonomous Vehicle at On-Ramp Merging},
  author={Zhang, Changzhu and Dai, Lifei and Zhang, Hao and Wang, Zhuping},
  journal={IEEE Transactions on Intelligent Transportation Systems},
  year={2025},
  publisher={IEEE}
}

@article{huang2023conditional,
  title={Conditional predictive behavior planning with inverse reinforcement learning for human-like autonomous driving},
  author={Huang, Zhiyu and Liu, Haochen and Wu, Jingda and Lv, Chen},
  journal={IEEE Transactions on Intelligent Transportation Systems},
  volume={24},
  number={7},
  pages={7244--7258},
  year={2023},
  publisher={IEEE}
}

@article{shi2024task,
  title={Task-Driven Autonomous Driving: Balanced Strategies Integrating Curriculum Reinforcement Learning and Residual Policy},
  author={Shi, Jiamin and Zhang, Tangyike and Zong, Ziqi and Chen, Shitao and Xin, Jingmin and Zheng, Nanning},
  journal={IEEE Robotics and Automation Letters},
  year={2024},
  publisher={IEEE}
}

@article{tang2023survey,
  title={A survey on automated driving system testing: Landscapes and trends},
  author={Tang, Shuncheng and Zhang, Zhenya and Zhang, Yi and Zhou, Jixiang and Guo, Yan and Liu, Shuang and Guo, Shengjian and Li, Yan-Fu and Ma, Lei and Xue, Yinxing and others},
  journal={ACM Transactions on Software Engineering and Methodology},
  volume={32},
  number={5},
  pages={1--62},
  year={2023},
  publisher={ACM New York, NY}
}

@inproceedings{lou2022testing,
  title={Testing of autonomous driving systems: where are we and where should we go?},
  author={Lou, Guannan and Deng, Yao and Zheng, Xi and Zhang, Mengshi and Zhang, Tianyi},
  booktitle={Proceedings of the 30th ACM Joint European Software Engineering Conference and Symposium on the Foundations of Software Engineering},
  pages={31--43},
  year={2022}
}

@article{chen2023milestones,
  title={Milestones in autonomous driving and intelligent vehicles—Part I: Control, computing system design, communication, HD map, testing, and human behaviors},
  author={Chen, Long and Li, Yuchen and Huang, Chao and Xing, Yang and Tian, Daxin and Li, Li and Hu, Zhongxu and Teng, Siyu and Lv, Chen and Wang, Jinjun and others},
  journal={IEEE Transactions on Systems, Man, and Cybernetics: Systems},
  volume={53},
  number={9},
  pages={5831--5847},
  year={2023},
  publisher={IEEE}
}

@inproceedings{wang2022adept,
  title={ADEPT: A testing platform for simulated autonomous driving},
  author={Wang, Sen and Sheng, Zhuheng and Xu, Jingwei and Chen, Taolue and Zhu, Junjun and Zhang, Shuhui and Yao, Yuan and Ma, Xiaoxing},
  booktitle={Proceedings of the 37th IEEE/ACM International Conference on Automated Software Engineering},
  pages={1--4},
  year={2022}
}

@inproceedings{rong2020lgsvl,
  title={Lgsvl simulator: A high fidelity simulator for autonomous driving},
  author={Rong, Guodong and Shin, Byung Hyun and Tabatabaee, Hadi and Lu, Qiang and Lemke, Steve and Mo{\v{z}}eiko, M{\=a}rti{\c{n}}{\v{s}} and Boise, Eric and Uhm, Geehoon and Gerow, Mark and Mehta, Shalin and others},
  booktitle={2020 IEEE 23rd International conference on intelligent transportation systems (ITSC)},
  pages={1--6},
  year={2020},
  organization={IEEE}
}

@misc{NVIDIAAutonomousVehicleSimulation,
  title = {{{NVIDIAAutonomous Vehicle Simulation}}},
  journal = {NVIDIA Developer},
  urldate = {2025-09-28},
  howpublished = {https://developer.nvidia.com/drive/simulation},
  langid = {english}
}

@article{li2022metadrive,
  title={Metadrive: Composing diverse driving scenarios for generalizable reinforcement learning},
  author={Li, Quanyi and Peng, Zhenghao and Feng, Lan and Zhang, Qihang and Xue, Zhenghai and Zhou, Bolei},
  journal={IEEE transactions on pattern analysis and machine intelligence},
  volume={45},
  number={3},
  pages={3461--3475},
  year={2022},
  publisher={IEEE}
}

@article{zheng2023opencda,
  title={OpenCDA-ROS: Enabling seamless integration of simulation and real-world cooperative driving automation},
  author={Zheng, Zhaoliang and Han, Xu and Xia, Xin and Gao, Letian and Xiang, Hao and Ma, Jiaqi},
  journal={IEEE Transactions on Intelligent Vehicles},
  volume={8},
  number={7},
  pages={3775--3780},
  year={2023},
  publisher={IEEE}
}

@article{hu2023simulation,
  title={How simulation helps autonomous driving: A survey of sim2real, digital twins, and parallel intelligence},
  author={Hu, Xuemin and Li, Shen and Huang, Tingyu and Tang, Bo and Huai, Rouxing and Chen, Long},
  journal={IEEE Transactions on Intelligent Vehicles},
  volume={9},
  number={1},
  pages={593--612},
  year={2023},
  publisher={IEEE}
}

@article{zhou2024would,
  title={How would autonomous vehicles behave in real-world crash scenarios?},
  author={Zhou, Rui and Zhang, Guoqing and Huang, Helai and Wei, Zhiyuan and Zhou, Hanchu and Jin, Jieling and Chang, Fangrong and Chen, Jiguang},
  journal={Accident Analysis \& Prevention},
  volume={202},
  pages={107572},
  year={2024},
  publisher={Elsevier}
}

@article{rossolini2023real,
  title={On the real-world adversarial robustness of real-time semantic segmentation models for autonomous driving},
  author={Rossolini, Giulio and Nesti, Federico and D’amico, Gianluca and Nair, Saasha and Biondi, Alessandro and Buttazzo, Giorgio},
  journal={IEEE Transactions on Neural Networks and Learning Systems},
  year={2023},
  publisher={IEEE}
}

@article{lee2024study,
  title={A study on the driving performance analysis for autonomous vehicles through the real-road field operational test platform},
  author={Lee, Dong-Whan and Kim, Tae-Lim and Kwon, Seong-Jin},
  journal={International Journal of Precision Engineering and Manufacturing},
  volume={25},
  number={6},
  pages={1231--1241},
  year={2024},
  publisher={Springer}
}

@article{shoukat2024autonomous,
  title={Autonomous driving test system under hybrid reality: The role of digital twin technology},
  author={Shoukat, Muhammad Usman and Yan, Lirong and Yan, Yukai and Zhang, Fan and Zhai, Yikang and Han, Peng and Nawaz, Saqib Ali and Raza, Muhammad Ahmad and Akbar, Muhammad Waqas and Hussain, Abid},
  journal={Internet of Things},
  volume={27},
  pages={101301},
  year={2024},
  publisher={Elsevier}
}

@article{argui2024advancements,
  title={Advancements in mixed reality for autonomous vehicle testing and advanced driver assistance systems: A survey},
  author={Argui, Imane and Gueriau, Maxime and Ainouz, Samia},
  journal={IEEE Transactions on Intelligent Transportation Systems},
  year={2024},
  publisher={IEEE}
}

@inproceedings{li2024ego,
  title={Is ego status all you need for open-loop end-to-end autonomous driving?},
  author={Li, Zhiqi and Yu, Zhiding and Lan, Shiyi and Li, Jiahan and Kautz, Jan and Lu, Tong and Alvarez, Jose M},
  booktitle={Proceedings of the IEEE/CVF Conference on Computer Vision and Pattern Recognition},
  pages={14864--14873},
  year={2024}
}

@article{sharath2021literature,
  title={A literature review of performance metrics of automated driving systems for on-road vehicles},
  author={Sharath, Mysore Narasimhamurthy and Mehran, Babak},
  journal={Frontiers in Future Transportation},
  volume={2},
  pages={759125},
  year={2021},
  publisher={Frontiers Media SA}
}

@article{li2023toward,
  title={Toward fair and thrilling autonomous racing: Governance rules and performance metrics for the autonomous one},
  author={Li, Bai and Fang, Yile and Xu, Tian'ao and Ma, Siji and Wang, Haorui and Wang, Yazhou and Li, Xinyuan and Zhang, Tantan and Bian, Xuepeng and Wang, Fei-Yue},
  journal={IEEE Transactions on Intelligent Vehicles},
  volume={8},
  number={8},
  pages={3974--3982},
  year={2023},
  publisher={IEEE}
}

@article{xu2022safebench,
  title={Safebench: A benchmarking platform for safety evaluation of autonomous vehicles},
  author={Xu, Chejian and Ding, Wenhao and Lyu, Weijie and Liu, Zuxin and Wang, Shuai and He, Yihan and Hu, Hanjiang and Zhao, Ding and Li, Bo},
  journal={Advances in Neural Information Processing Systems},
  volume={35},
  pages={25667--25682},
  year={2022}
}

@inproceedings{inoue2024nuscenes,
  title={Nuscenes-mqa: Integrated evaluation of captions and qa for autonomous driving datasets using markup annotations},
  author={Inoue, Yuichi and Yada, Yuki and Tanahashi, Kotaro and Yamaguchi, Yu},
  booktitle={Proceedings of the IEEE/CVF Winter Conference on Applications of Computer Vision},
  pages={930--938},
  year={2024}
}

@article{torabi2018behavioral,
  title={Behavioral cloning from observation},
  author={Torabi, Faraz and Warnell, Garrett and Stone, Peter},
  journal={arXiv preprint arXiv:1805.01954},
  year={2018}
}

@inproceedings{florence2022implicit,
  title={Implicit behavioral cloning},
  author={Florence, Pete and Lynch, Corey and Zeng, Andy and Ramirez, Oscar A and Wahid, Ayzaan and Downs, Laura and Wong, Adrian and Lee, Johnny and Mordatch, Igor and Tompson, Jonathan},
  booktitle={Conference on robot learning},
  pages={158--168},
  year={2022},
  organization={PMLR}
}

@inproceedings{codevilla2019exploring,
  title={Exploring the limitations of behavior cloning for autonomous driving},
  author={Codevilla, Felipe and Santana, Eder and L{\'o}pez, Antonio M and Gaidon, Adrien},
  booktitle={Proceedings of the IEEE/CVF international conference on computer vision},
  pages={9329--9338},
  year={2019}
}

@article{adams2022survey,
  title={A survey of inverse reinforcement learning},
  author={Adams, Stephen and Cody, Tyler and Beling, Peter A},
  journal={Artificial Intelligence Review},
  volume={55},
  number={6},
  pages={4307--4346},
  year={2022},
  publisher={Springer}
}

@article{couto2024hierarchical,
  author={Couto, Gustavo Claudio Karl and Antonelo, Eric Aislan},
  journal={IEEE Transactions on Intelligent Vehicles}, 
  title={Hierarchical Generative Adversarial Imitation Learning With Mid-Level Input Generation for Autonomous Driving on Urban Environments}, 
  year={2025},
  volume={10},
  number={3},
  pages={1841-1854},
  doi={10.1109/TIV.2024.3436587}}

@article{chang2021mitigating,
  title={Mitigating covariate shift in imitation learning via offline data with partial coverage},
  author={Chang, Jonathan and Uehara, Masatoshi and Sreenivas, Dhruv and Kidambi, Rahul and Sun, Wen},
  journal={Advances in Neural Information Processing Systems},
  volume={34},
  pages={965--979},
  year={2021}
}

@article{seo2024mitigating,
  title={Mitigating covariate shift in behavioral cloning via robust stationary distribution correction},
  author={Seo, Seokin and Lee, Byung-Jun and Lee, Jongmin and Hwang, HyeongJoo and Yang, Hongseok and Kim, Kee-Eung},
  journal={Advances in Neural Information Processing Systems},
  volume={37},
  pages={109177--109201},
  year={2024}
}

@article{popov2024mitigating,
  title={Mitigating covariate shift in imitation learning for autonomous vehicles using latent space generative world models},
  author={Popov, Alexander and Degirmenci, Alperen and Wehr, David and Hegde, Shashank and Oldja, Ryan and Kamenev, Alexey and Douillard, Bertrand and Nist{\'e}r, David and Muller, Urs and Bhargava, Ruchi and others},
  journal={arXiv preprint arXiv:2409.16663},
  year={2024}
}

@inproceedings{hallgarten2024can,
  title={Can vehicle motion planning generalize to realistic long-tail scenarios?},
  author={Hallgarten, Marcel and Zapata, Julian and Stoll, Martin and Renz, Katrin and Zell, Andreas},
  booktitle={2024 IEEE/RSJ International Conference on Intelligent Robots and Systems (IROS)},
  pages={5388--5395},
  year={2024},
  organization={IEEE}
}

@article{zhang2025systematic,
  title={A systematic review on long-tailed learning},
  author={Zhang, Chongsheng and Almpanidis, George and Fan, Gaojuan and Deng, Binquan and Zhang, Yanbo and Liu, Ji and Kamel, Aouaidjia and Soda, Paolo and Gama, Jo{\~a}o},
  journal={IEEE Transactions on Neural Networks and Learning Systems},
  year={2025},
  publisher={IEEE}
}

@inproceedings{ma2025novel,
  title={Novel view synthesis under large-deviation viewpoint for autonomous driving},
  author={Ma, Xin and Zhang, Jiguang and Lu, Peng and Xu, Shibiao and Pan, Chengwei},
  booktitle={Proceedings of the AAAI Conference on Artificial Intelligence},
  volume={39},
  number={6},
  pages={6000--6008},
  year={2025}
}

@inproceedings{pan2025transfer,
  title={Transfer Your Perspective: Controllable 3D Generation from Any Viewpoint in a Driving Scene},
  author={Pan, Tai-Yu and Jeon, Sooyoung and Fan, Mengdi and Yoo, Jinsu and Feng, Zhenyang and Campbell, Mark and Weinberger, Kilian Q and Hariharan, Bharath and Chao, Wei-Lun},
  booktitle={Proceedings of the Computer Vision and Pattern Recognition Conference},
  pages={12027--12036},
  year={2025}
}

@inproceedings{yu2025enduring,
  title={Enduring, Efficient and Robust Trajectory Prediction Attack in Autonomous Driving via Optimization-Driven Multi-Frame Perturbation Framework},
  author={Yu, Yi and Han, Weizhen and Wu, Libing and Liu, Bingyi and Wang, Enshu and Zhang, Zhuangzhuang},
  booktitle={Proceedings of the Computer Vision and Pattern Recognition Conference},
  pages={17229--17238},
  year={2025}
}

@article{wang2024counterfactual,
  title={Counterfactual explanations for deep learning-based traffic forecasting},
  author={Wang, Rushan and Xin, Yanan and Zhang, Yatao and Perez-Cruz, Fernando and Raubal, Martin},
  journal={arXiv preprint arXiv:2405.00456},
  year={2024}
}

@article{ye2023counterfactual,
  title={Counterfactual evolutionary reasoning for virtual driver reinforcement learning in safe driving},
  author={Ye, Peijun and Qi, Hao and Zhu, Fenghua and Lv, Yisheng},
  journal={IEEE Transactions on Intelligent Vehicles},
  volume={8},
  number={12},
  pages={4696--4705},
  year={2023},
  publisher={IEEE}
}

@inproceedings{ross2011reduction,
  title={A reduction of imitation learning and structured prediction to no-regret online learning},
  author={Ross, St{\'e}phane and Gordon, Geoffrey and Bagnell, Drew},
  booktitle={Proceedings of the fourteenth international conference on artificial intelligence and statistics},
  pages={627--635},
  year={2011},
  organization={JMLR Workshop and Conference Proceedings}
}

@inproceedings{bicer2019sample,
  title={Sample efficient interactive end-to-end deep learning for self-driving cars with selective multi-class safe dataset aggregation},
  author={Bicer, Yunus and Alizadeh, Ali and Ure, Nazim Kemal and Erdogan, Ahmetcan and Kizilirmak, Orkun},
  booktitle={2019 IEEE/RSJ International Conference on Intelligent Robots and Systems (IROS)},
  pages={2629--2634},
  year={2019},
  organization={IEEE}
}

@article{ahn2022autonomous,
  title={Autonomous driving using imitation learning with look ahead point for semi structured environments},
  author={Ahn, Joonwoo and Kim, Minsoo and Park, Jaeheung},
  journal={Scientific Reports},
  volume={12},
  number={1},
  pages={21285},
  year={2022},
  publisher={Nature Publishing Group UK London}
}

@inproceedings{tang2023personalized,
  title={Personalized decision-making and control for automated vehicles based on generative adversarial imitation learning},
  author={Tang, Xinyue and Yuan, Kang and Li, Shangwen and Yang, Shuo and Zhou, Zewei and Huang, Yanjun},
  booktitle={2023 IEEE 26th International Conference on Intelligent Transportation Systems (ITSC)},
  pages={4806--4812},
  year={2023},
  organization={IEEE}
}

@article{cheng2024pluto,
  title={Pluto: Pushing the limit of imitation learning-based planning for autonomous driving},
  author={Cheng, Jie and Chen, Yingbing and Chen, Qifeng},
  journal={arXiv preprint arXiv:2404.14327},
  year={2024}
}

@article{shakya2023reinforcement,
  title={Reinforcement learning algorithms: A brief survey},
  author={Shakya, Ashish Kumar and Pillai, Gopinatha and Chakrabarty, Sohom},
  journal={Expert Systems with Applications},
  volume={231},
  pages={120495},
  year={2023},
  publisher={Elsevier}
}

@article{matsuo2022deep,
  title={Deep learning, reinforcement learning, and world models},
  author={Matsuo, Yutaka and LeCun, Yann and Sahani, Maneesh and Precup, Doina and Silver, David and Sugiyama, Masashi and Uchibe, Eiji and Morimoto, Jun},
  journal={Neural Networks},
  volume={152},
  pages={267--275},
  year={2022},
  publisher={Elsevier}
}

@article{zhao2024survey,
  title={A survey on recent advancements in autonomous driving using deep reinforcement learning: Applications, challenges, and solutions},
  author={Zhao, Rui and Li, Yun and Fan, Yuze and Gao, Fei and Tsukada, Manabu and Gao, Zhenhai},
  journal={IEEE Transactions on Intelligent Transportation Systems},
  year={2024},
  publisher={IEEE}
}

@inproceedings{li2024think2drive,
  title={Think2drive: Efficient reinforcement learning by thinking with latent world model for autonomous driving (in carla-v2)},
  author={Li, Qifeng and Jia, Xiaosong and Wang, Shaobo and Yan, Junchi},
  booktitle={European Conference on Computer Vision},
  pages={142--158},
  year={2024},
  organization={Springer}
}

@article{mo2021safe,
  title={Safe reinforcement learning for autonomous vehicle using monte carlo tree search},
  author={Mo, Shuojie and Pei, Xiaofei and Wu, Chaoxian},
  journal={IEEE Transactions on Intelligent Transportation Systems},
  volume={23},
  number={7},
  pages={6766--6773},
  year={2021},
  publisher={IEEE}
}

@article{xu2026towards,
  title={Towards Safe and Robust Autonomous Vehicle Platooning: A Self-Organizing Cooperative Decision-making Framework},
  author={Xu, Chengkai and Deng, Zihao and Liu, Jiaqi and Kong, Aijing and Tang, Yu and Huang, Chao and Hang, Peng},
  journal={IEEE Transactions on Vehicular Technology},
  year={2026},
  publisher={IEEE}
}

@article{cai2021decision,
  title={A decision control method for autonomous driving based on multi-task reinforcement learning},
  author={Cai, Yingfeng and Yang, Shaoqing and Wang, Hai and Teng, Chenglong and Chen, Long},
  journal={IEEE Access},
  volume={9},
  pages={154553--154562},
  year={2021},
  publisher={IEEE}
}

@article{akhauri2021improving,
  title={Improving generalization of transfer learning across domains using spatio-temporal features in autonomous driving},
  author={Akhauri, Shivam and Zheng, Laura and Goldstein, Tom and Lin, Ming},
  journal={arXiv preprint arXiv:2103.08116},
  year={2021}
}

@article{lee2024imagination,
  title={Imagination-augmented hierarchical reinforcement learning for safe and interactive autonomous driving in urban environments},
  author={Lee, Sang-Hyun and Jung, Yoonjae and Seo, Seung-Woo},
  journal={IEEE Transactions on Intelligent Transportation Systems},
  year={2024},
  publisher={IEEE}
}

@article{hu2024long,
  title={Long and short-term constraints driven safe reinforcement learning for autonomous driving},
  author={Hu, Xuemin and Chen, Pan and Wen, Yijun and Tang, Bo and Chen, Long},
  journal={arXiv preprint arXiv:2403.18209},
  year={2024}
}

@article{he2023toward,
  title={Toward personalized decision making for autonomous vehicles: a constrained multi-objective reinforcement learning technique},
  author={He, Xiangkun and Lv, Chen},
  journal={Transportation research part C: emerging technologies},
  volume={156},
  pages={104352},
  year={2023},
  publisher={Elsevier}
}

@article{zhang2024cvar,
  title={CVaR-constrained policy optimization for safe reinforcement learning},
  author={Zhang, Qiyuan and Leng, Shu and Ma, Xiaoteng and Liu, Qihan and Wang, Xueqian and Liang, Bin and Liu, Yu and Yang, Jun},
  journal={IEEE transactions on neural networks and learning systems},
  volume={36},
  number={1},
  pages={830--841},
  year={2024},
  publisher={IEEE}
}

@article{tang2023uncertainty,
  title={Uncertainty-aware decision-making for autonomous driving at uncontrolled intersections},
  author={Tang, Xiaolin and Zhong, Guichuan and Li, Shen and Yang, Kai and Shu, Keqi and Cao, Dongpu and Lin, Xianke},
  journal={IEEE Transactions on Intelligent Transportation Systems},
  volume={24},
  number={9},
  pages={9725--9735},
  year={2023},
  publisher={IEEE}
}

@article{gangopadhyay2022safe,
  title={Safe and stable RL (S 2 RL) driving policies using control barrier and control Lyapunov functions},
  author={Gangopadhyay, Briti and Dasgupta, Pallab and Dey, Soumyajit},
  journal={IEEE Transactions on Intelligent Vehicles},
  volume={8},
  number={2},
  pages={1889--1899},
  year={2022},
  publisher={IEEE}
}

@article{zhuang2024hgrl,
  title={HGRL: Human-driving-data guided reinforcement learning for autonomous driving},
  author={Zhuang, Hejian and Chu, Hongqing and Wang, Yafei and Gao, Bingzhao and Chen, Hong},
  journal={IEEE Transactions on Intelligent Vehicles},
  year={2024},
  publisher={IEEE}
}

@article{yang2025human,
  title={Human-guided continual learning for personalized decision-making of autonomous driving},
  author={Yang, Haohan and Zhou, Yanxin and Wu, Jingda and Liu, Haochen and Yang, Lie and Lv, Chen},
  journal={IEEE Transactions on Intelligent Transportation Systems},
  year={2025},
  publisher={IEEE}
}

@article{anzalone2022end,
  title={An end-to-end curriculum learning approach for autonomous driving scenarios},
  author={Anzalone, Luca and Barra, Paola and Barra, Silvio and Castiglione, Aniello and Nappi, Michele},
  journal={IEEE Transactions on Intelligent Transportation Systems},
  volume={23},
  number={10},
  pages={19817--19826},
  year={2022},
  publisher={IEEE}
}

@article{hong2025effective,
  title={Effective Learning Mechanism Based on Reward-Oriented Hierarchies for Sim-to-Real Adaption in Autonomous Driving Systems},
  author={Hong, Zhiming},
  journal={IEEE Transactions on Intelligent Transportation Systems},
  year={2025},
  publisher={IEEE}
}

@article{ho2020denoising,
  title={Denoising diffusion probabilistic models},
  author={Ho, Jonathan and Jain, Ajay and Abbeel, Pieter},
  journal={Advances in neural information processing systems},
  volume={33},
  pages={6840--6851},
  year={2020}
}

@article{chi2023diffusion,
  title={Diffusion policy: Visuomotor policy learning via action diffusion},
  author={Chi, Cheng and Xu, Zhenjia and Feng, Siyuan and Cousineau, Eric and Du, Yilun and Burchfiel, Benjamin and Tedrake, Russ and Song, Shuran},
  journal={The International Journal of Robotics Research},
  pages={02783649241273668},
  year={2023},
  publisher={SAGE Publications Sage UK: London, England}
}

@article{ma2025efficient,
  title={Efficient diffusion models: A comprehensive survey from principles to practices},
  author={Ma, Zhiyuan and Zhang, Yuzhu and Jia, Guoli and Zhao, Liangliang and Ma, Yichao and Ma, Mingjie and Liu, Gaofeng and Zhang, Kaiyan and Ding, Ning and Li, Jianjun and others},
  journal={IEEE Transactions on Pattern Analysis and Machine Intelligence},
  year={2025},
  publisher={IEEE}
}

@article{peng2025diffusion,
  title={Diffusion models for intelligent transportation systems: A survey},
  author={Peng, Mingxing and Chen, Kehua and Guo, Xusen and Zhang, Qiming and Zhong, Hui and Zhu, Meixin and Yang, Hai},
  journal={IEEE Transactions on Intelligent Transportation Systems},
  year={2025},
  publisher={IEEE}
}

@article{li2024learning,
  title={Learning multimodal behaviors from scratch with diffusion policy gradient},
  author={Li, Steven and Krohn, Rickmer and Chen, Tao and Ajay, Anurag and Agrawal, Pulkit and Chalvatzaki, Georgia},
  journal={Advances in Neural Information Processing Systems},
  volume={37},
  pages={38456--38479},
  year={2024}
}

@inproceedings{liao2025diffusiondrive,
  title={Diffusiondrive: Truncated diffusion model for end-to-end autonomous driving},
  author={Liao, Bencheng and Chen, Shaoyu and Yin, Haoran and Jiang, Bo and Wang, Cheng and Yan, Sixu and Zhang, Xinbang and Li, Xiangyu and Zhang, Ying and Zhang, Qian and others},
  booktitle={Proceedings of the Computer Vision and Pattern Recognition Conference},
  pages={12037--12047},
  year={2025}
}

@article{jiang2025diffvla,
  title={Diffvla: Vision-language guided diffusion planning for autonomous driving},
  author={Jiang, Anqing and Gao, Yu and Sun, Zhigang and Wang, Yiru and Wang, Jijun and Chai, Jinghao and Cao, Qian and Heng, Yuweng and Jiang, Hao and Dong, Yunda and others},
  journal={arXiv preprint arXiv:2505.19381},
  year={2025}
}

@article{zheng2025diffusion,
  title={Diffusion-based planning for autonomous driving with flexible guidance},
  author={Zheng, Yinan and Liang, Ruiming and Zheng, Kexin and Zheng, Jinliang and Mao, Liyuan and Li, Jianxiong and Gu, Weihao and Ai, Rui and Li, Shengbo Eben and Zhan, Xianyuan and others},
  journal={arXiv preprint arXiv:2501.15564},
  year={2025}
}

@inproceedings{huang2025gen,
  title={Gen-drive: Enhancing diffusion generative driving policies with reward modeling and reinforcement learning fine-tuning},
  author={Huang, Zhiyu and Weng, Xinshuo and Igl, Maximilian and Chen, Yuxiao and Cao, Yulong and Ivanovic, Boris and Pavone, Marco and Lv, Chen},
  booktitle={2025 IEEE International Conference on Robotics and Automation (ICRA)},
  pages={3445--3451},
  year={2025},
  organization={IEEE}
}

@article{zhou2023adaptive,
  title={Adaptive online replanning with diffusion models},
  author={Zhou, Siyuan and Du, Yilun and Zhang, Shun and Xu, Mengdi and Shen, Yikang and Xiao, Wei and Yeung, Dit-Yan and Gan, Chuang},
  journal={Advances in Neural Information Processing Systems},
  volume={36},
  pages={44000--44016},
  year={2023}
}

@article{carvalho2025motion,
  title={Motion planning diffusion: Learning and adapting robot motion planning with diffusion models},
  author={Carvalho, Joao and Le, An T and Kicki, Piotr and Koert, Dorothea and Peters, Jan},
  journal={IEEE Transactions on Robotics},
  year={2025},
  publisher={IEEE}
}

@article{sun2023conformal,
  title={Conformal prediction for uncertainty-aware planning with diffusion dynamics model},
  author={Sun, Jiankai and Jiang, Yiqi and Qiu, Jianing and Nobel, Parth and Kochenderfer, Mykel J and Schwager, Mac},
  journal={Advances in Neural Information Processing Systems},
  volume={36},
  pages={80324--80337},
  year={2023}
}

@article{wu2024dexterous,
  title={Dexterous Functional Pre-Grasp Manipulation with Diffusion Policy},
  author={Wu, Tianhao and Gan, Yunchong and Wu, Mingdong and Cheng, Jingbo and Yang, Yaodong and Zhu, Yixin and Dong, Hao},
  journal={arXiv preprint arXiv:2403.12421},
  year={2024}
}

@article{jiang2025adaptive,
  title={Adaptive Diffusion Policy Optimization for Robotic Manipulation},
  author={Jiang, Huiyun and Yang, Zhuang},
  journal={arXiv preprint arXiv:2505.08376},
  year={2025}
}

@article{kang2023efficient,
  title={Efficient diffusion policies for offline reinforcement learning},
  author={Kang, Bingyi and Ma, Xiao and Du, Chao and Pang, Tianyu and Yan, Shuicheng},
  journal={Advances in Neural Information Processing Systems},
  volume={36},
  pages={67195--67212},
  year={2023}
}

@inproceedings{liang2024skilldiffuser,
  title={Skilldiffuser: Interpretable hierarchical planning via skill abstractions in diffusion-based task execution},
  author={Liang, Zhixuan and Mu, Yao and Ma, Hengbo and Tomizuka, Masayoshi and Ding, Mingyu and Luo, Ping},
  booktitle={Proceedings of the IEEE/CVF Conference on Computer Vision and Pattern Recognition},
  pages={16467--16476},
  year={2024}
}

@article{hou2024diffusion,
  title={Diffusion transformer policy},
  author={Hou, Zhi and Zhang, Tianyi and Xiong, Yuwen and Pu, Hengjun and Zhao, Chengyang and Tong, Ronglei and Qiao, Yu and Dai, Jifeng and Chen, Yuntao},
  journal={arXiv preprint arXiv:2410.15959},
  year={2024}
}

@article{wintel2025using,
  title={Using Diffusion Ensembles to Estimate Uncertainty for End-to-End Autonomous Driving},
  author={Wintel, Florian and H{\o}eg, Sigmund H and Kiss, Gabriel and Lindseth, Frank},
  journal={arXiv preprint arXiv:2506.00560},
  year={2025}
}

@inproceedings{jiang2025survey,
  title={A survey on vision-language-action models for autonomous driving},
  author={Jiang, Sicong and Huang, Zilin and Qian, Kangan and Luo, Ziang and Zhu, Tianze and Zhong, Yang and Tang, Yihong and Kong, Menglin and Wang, Yunlong and Jiao, Siwen and others},
  booktitle={Proceedings of the IEEE/CVF International Conference on Computer Vision},
  pages={4524--4536},
  year={2025}
}

@article{yin2024survey,
  title={A survey on multimodal large language models},
  author={Yin, Shukang and Fu, Chaoyou and Zhao, Sirui and Li, Ke and Sun, Xing and Xu, Tong and Chen, Enhong},
  journal={National Science Review},
  volume={11},
  number={12},
  pages={nwae403},
  year={2024},
  publisher={Oxford University Press}
}

@article{ma2024survey,
  title={A survey on vision-language-action models for embodied ai},
  author={Ma, Yueen and Song, Zixing and Zhuang, Yuzheng and Hao, Jianye and King, Irwin},
  journal={arXiv preprint arXiv:2405.14093},
  year={2024}
}

@inproceedings{arai2025covla,
  title={Covla: Comprehensive vision-language-action dataset for autonomous driving},
  author={Arai, Hidehisa and Miwa, Keita and Sasaki, Kento and Watanabe, Kohei and Yamaguchi, Yu and Aoki, Shunsuke and Yamamoto, Issei},
  booktitle={2025 IEEE/CVF Winter Conference on Applications of Computer Vision (WACV)},
  pages={1933--1943},
  year={2025},
  organization={IEEE}
}

@inproceedings{marcu2024lingoqa,
  title={Lingoqa: Visual question answering for autonomous driving},
  author={Marcu, Ana-Maria and Chen, Long and H{\"u}nermann, Jan and Karnsund, Alice and Hanotte, Benoit and Chidananda, Prajwal and Nair, Saurabh and Badrinarayanan, Vijay and Kendall, Alex and Shotton, Jamie and others},
  booktitle={European Conference on Computer Vision},
  pages={252--269},
  year={2024},
  organization={Springer}
}

@article{zhang2024wisead,
  title={Wisead: Knowledge augmented end-to-end autonomous driving with vision-language model},
  author={Zhang, Songyan and Huang, Wenhui and Gao, Zihui and Chen, Hao and Lv, Chen},
  journal={arXiv preprint arXiv:2412.09951},
  year={2024}
}

@article{long2024vlm,
  title={Vlm-mpc: Vision language foundation model (vlm)-guided model predictive controller (mpc) for autonomous driving},
  author={Long, Keke and Shi, Haotian and Liu, Jiaxi and Li, Xiaopeng},
  journal={arXiv preprint arXiv:2408.04821},
  year={2024}
}

@article{chi2025impromptu,
  title={Impromptu VLA: Open Weights and Open Data for Driving Vision-Language-Action Models},
  author={Chi, Haohan and Gao, Huan-ang and Liu, Ziming and Liu, Jianing and Liu, Chenyu and Li, Jinwei and Yang, Kaisen and Yu, Yangcheng and Wang, Zeda and Li, Wenyi and others},
  journal={arXiv preprint arXiv:2505.23757},
  year={2025}
}

@inproceedings{pan2024vlp,
  title={Vlp: Vision language planning for autonomous driving},
  author={Pan, Chenbin and Yaman, Burhaneddin and Nesti, Tommaso and Mallik, Abhirup and Allievi, Alessandro G and Velipasalar, Senem and Ren, Liu},
  booktitle={Proceedings of the IEEE/CVF Conference on Computer Vision and Pattern Recognition},
  pages={14760--14769},
  year={2024}
}

@article{zhou2025autovla,
  title={AutoVLA: A Vision-Language-Action Model for End-to-End Autonomous Driving with Adaptive Reasoning and Reinforcement Fine-Tuning},
  author={Zhou, Zewei and Cai, Tianhui and Zhao, Seth Z and Zhang, Yun and Huang, Zhiyu and Zhou, Bolei and Ma, Jiaqi},
  journal={arXiv preprint arXiv:2506.13757},
  year={2025}
}

@article{li2025recogdrive,
  title={ReCogDrive: A Reinforced Cognitive Framework for End-to-End Autonomous Driving},
  author={Li, Yongkang and Xiong, Kaixin and Guo, Xiangyu and Li, Fang and Yan, Sixu and Xu, Gangwei and Zhou, Lijun and Chen, Long and Sun, Haiyang and Wang, Bing and others},
  journal={arXiv preprint arXiv:2506.08052},
  year={2025}
}

@article{jiang2025structured,
  title={Structured Labeling Enables Faster Vision-Language Models for End-to-End Autonomous Driving},
  author={Jiang, Hao and Hu, Chuan and Shi, Yukang and He, Yuan and Wang, Ke and Zhang, Xi and Zhang, Zhipeng},
  journal={arXiv preprint arXiv:2506.05442},
  year={2025}
}

@article{li2024womd,
  title={WOMD-Reasoning: A Large-Scale Dataset for Interaction Reasoning in Driving},
  author={Li, Yiheng and Fan, Cunxin and Ge, Chongjian and Zhao, Zhihao and Li, Chenran and Xu, Chenfeng and Yao, Huaxiu and Tomizuka, Masayoshi and Zhou, Bolei and Tang, Chen and others},
  journal={arXiv preprint arXiv:2407.04281},
  year={2024}
}

@article{zeng2025futuresightdrive,
  title={FutureSightDrive: Thinking Visually with Spatio-Temporal CoT for Autonomous Driving},
  author={Zeng, Shuang and Chang, Xinyuan and Xie, Mengwei and Liu, Xinran and Bai, Yifan and Pan, Zheng and Xu, Mu and Wei, Xing},
  journal={arXiv preprint arXiv:2505.17685},
  year={2025}
}

@article{qu2025vl,
  title={VL-SAFE: Vision-Language Guided Safety-Aware Reinforcement Learning with World Models for Autonomous Driving},
  author={Qu, Yansong and Huang, Zilin and Sheng, Zihao and Chen, Jiancong and Chen, Sikai and Labi, Samuel},
  journal={arXiv preprint arXiv:2505.16377},
  year={2025}
}

@article{qian2025agentthink,
  title={Agentthink: A unified framework for tool-augmented chain-of-thought reasoning in vision-language models for autonomous driving},
  author={Qian, Kangan and Jiang, Sicong and Zhong, Yang and Luo, Ziang and Huang, Zilin and Zhu, Tianze and Jiang, Kun and Yang, Mengmeng and Fu, Zheng and Miao, Jinyu and others},
  journal={arXiv preprint arXiv:2505.15298},
  year={2025}
}

@article{liu2025dsdrive,
  title={DSDrive: Distilling Large Language Model for Lightweight End-to-End Autonomous Driving with Unified Reasoning and Planning},
  author={Liu, Wenru and Liu, Pei and Ma, Jun},
  journal={arXiv preprint arXiv:2505.05360},
  year={2025}
}

@article{baumann2025enhancing,
  title={Enhancing autonomous driving systems with on-board deployed large language models},
  author={Baumann, Nicolas and Hu, Cheng and Sivasothilingam, Paviththiren and Qin, Haotong and Xie, Lei and Magno, Michele and Benini, Luca},
  journal={arXiv preprint arXiv:2504.11514},
  year={2025}
}

@article{liao2025cot,
  title={Cot-drive: Efficient motion forecasting for autonomous driving with llms and chain-of-thought prompting},
  author={Liao, Haicheng and Kong, Hanlin and Wang, Bonan and Wang, Chengyue and Ye, Wang and He, Zhengbing and Xu, Chengzhong and Li, Zhenning},
  journal={IEEE Transactions on Artificial Intelligence},
  year={2025},
  publisher={IEEE}
}

@article{zhang2025evaluation,
  title={Evaluation of safety cognition capability in vision-language models for autonomous driving},
  author={Zhang, Enming and Gong, Peizhe and Dai, Xingyuan and Huang, Min and Lv, Yisheng and Miao, Qinghai},
  journal={arXiv preprint arXiv:2503.06497},
  year={2025}
}

@inproceedings{hegde2025distilling,
  title={Distilling multi-modal large language models for autonomous driving},
  author={Hegde, Deepti and Yasarla, Rajeev and Cai, Hong and Han, Shizhong and Bhattacharyya, Apratim and Mahajan, Shweta and Liu, Litian and Garrepalli, Risheek and Patel, Vishal M and Porikli, Fatih},
  booktitle={Proceedings of the Computer Vision and Pattern Recognition Conference},
  pages={27575--27585},
  year={2025}
}

@article{xie2025vlms,
  title={Are vlms ready for autonomous driving? an empirical study from the reliability, data, and metric perspectives},
  author={Xie, Shaoyuan and Kong, Lingdong and Dong, Yuhao and Sima, Chonghao and Zhang, Wenwei and Chen, Qi Alfred and Liu, Ziwei and Pan, Liang},
  journal={arXiv preprint arXiv:2501.04003},
  year={2025}
}

@article{xing2024autotrust,
  title={Autotrust: Benchmarking trustworthiness in large vision language models for autonomous driving},
  author={Xing, Shuo and Hua, Hongyuan and Gao, Xiangbo and Zhu, Shenzhe and Li, Renjie and Tian, Kexin and Li, Xiaopeng and Huang, Heng and Yang, Tianbao and Wang, Zhangyang and others},
  journal={arXiv preprint arXiv:2412.15206},
  year={2024}
}

@inproceedings{ding2025hint,
  title={Hint-AD: Holistically Aligned Interpretability in End-to-End Autonomous Driving},
  author={Ding, Kairui and Chen, Boyuan and Su, Yuchen and Gao, Huan-ang and Jin, Bu and Sima, Chonghao and Li, Xiaohui and Zhang, Wuqiang and Barsch, Paul and Li, Hongyang and others},
  booktitle={Conference on Robot Learning},
  pages={3742--3765},
  year={2025},
  organization={PMLR}
}

@article{tian2024drivevlm,
  title={Drivevlm: The convergence of autonomous driving and large vision-language models},
  author={Tian, Xiaoyu and Gu, Junru and Li, Bailin and Liu, Yicheng and Wang, Yang and Zhao, Zhiyong and Zhan, Kun and Jia, Peng and Lang, Xianpeng and Zhao, Hang},
  journal={arXiv preprint arXiv:2402.12289},
  year={2024}
}

@article{friston2021world,
  title={World model learning and inference},
  author={Friston, Karl and Moran, Rosalyn J and Nagai, Yukie and Taniguchi, Tadahiro and Gomi, Hiroaki and Tenenbaum, Josh},
  journal={Neural Networks},
  volume={144},
  pages={573--590},
  year={2021},
  publisher={Elsevier}
}

@article{tu2025role,
  title={The role of world models in shaping autonomous driving: A comprehensive survey},
  author={Tu, Sifan and Zhou, Xin and Liang, Dingkang and Jiang, Xingyu and Zhang, Yumeng and Li, Xiaofan and Bai, Xiang},
  journal={arXiv preprint arXiv:2502.10498},
  year={2025}
}

@article{xie20252d,
  title={From 2D to 3D Cognition: A Brief Survey of General World Models},
  author={Xie, Ningwei and Tian, Zizi and Yang, Lei and Zhang, Xiao-Ping and Guo, Meng and Li, Jie},
  journal={arXiv preprint arXiv:2506.20134},
  year={2025}
}

@article{long2025survey,
  title={A Survey: Learning Embodied Intelligence from Physical Simulators and World Models},
  author={Long, Xiaoxiao and Zhao, Qingrui and Zhang, Kaiwen and Zhang, Zihao and Wang, Dingrui and Liu, Yumeng and Shu, Zhengjie and Lu, Yi and Wang, Shouzheng and Wei, Xinzhe and others},
  journal={arXiv preprint arXiv:2507.00917},
  year={2025}
}

@article{feng2025survey,
  title={A survey of world models for autonomous driving},
  author={Feng, Tuo and Wang, Wenguan and Yang, Yi},
  journal={arXiv preprint arXiv:2501.11260},
  year={2025}
}

@inproceedings{sekar2020planning,
  title={Planning to explore via self-supervised world models},
  author={Sekar, Ramanan and Rybkin, Oleh and Daniilidis, Kostas and Abbeel, Pieter and Hafner, Danijar and Pathak, Deepak},
  booktitle={International conference on machine learning},
  pages={8583--8592},
  year={2020},
  organization={PMLR}
}

@inproceedings{lin2020improving,
  title={Improving generative imagination in object-centric world models},
  author={Lin, Zhixuan and Wu, Yi-Fu and Peri, Skand and Fu, Bofeng and Jiang, Jindong and Ahn, Sungjin},
  booktitle={International conference on machine learning},
  pages={6140--6149},
  year={2020},
  organization={PMLR}
}

@article{ding2025understanding,
  title={Understanding world or predicting future? a comprehensive survey of world models},
  author={Ding, Jingtao and Zhang, Yunke and Shang, Yu and Zhang, Yuheng and Zong, Zefang and Feng, Jie and Yuan, Yuan and Su, Hongyuan and Li, Nian and Sukiennik, Nicholas and others},
  journal={ACM Computing Surveys},
  volume={58},
  number={3},
  pages={1--38},
  year={2025},
  publisher={ACM New York, NY}
}

@article{zhou2025hermes,
  title={Hermes: A unified self-driving world model for simultaneous 3d scene understanding and generation},
  author={Zhou, Xin and Liang, Dingkang and Tu, Sifan and Chen, Xiwu and Ding, Yikang and Zhang, Dingyuan and Tan, Feiyang and Zhao, Hengshuang and Bai, Xiang},
  journal={arXiv preprint arXiv:2501.14729},
  year={2025}
}

@article{gao2024vista,
  title={Vista: A generalizable driving world model with high fidelity and versatile controllability},
  author={Gao, Shenyuan and Yang, Jiazhi and Chen, Li and Chitta, Kashyap and Qiu, Yihang and Geiger, Andreas and Zhang, Jun and Li, Hongyang},
  journal={Advances in Neural Information Processing Systems},
  volume={37},
  pages={91560--91596},
  year={2024}
}

@article{gu2024dome,
  title={Dome: Taming diffusion model into high-fidelity controllable occupancy world model},
  author={Gu, Songen and Yin, Wei and Jin, Bu and Guo, Xiaoyang and Wang, Junming and Li, Haodong and Zhang, Qian and Long, Xiaoxiao},
  journal={arXiv preprint arXiv:2410.10429},
  year={2024}
}

@article{vafa2024evaluating,
  title={Evaluating the world model implicit in a generative model},
  author={Vafa, Keyon and Chen, Justin Y and Rambachan, Ashesh and Kleinberg, Jon and Mullainathan, Sendhil},
  journal={Advances in Neural Information Processing Systems},
  volume={37},
  pages={26941--26975},
  year={2024}
}

@article{kwon2025toward,
  title={Toward Stable World Models: Measuring and Addressing World Instability in Generative Environments},
  author={Kwon, Soonwoo and Kim, Jin-Young and Go, Hyojun and Baek, Kyungjune},
  journal={arXiv preprint arXiv:2503.08122},
  year={2025}
}

@inproceedings{min2024driveworld,
  title={Driveworld: 4d pre-trained scene understanding via world models for autonomous driving},
  author={Min, Chen and Zhao, Dawei and Xiao, Liang and Zhao, Jian and Xu, Xinli and Zhu, Zheng and Jin, Lei and Li, Jianshu and Guo, Yulan and Xing, Junliang and others},
  booktitle={Proceedings of the IEEE/CVF conference on computer vision and pattern recognition},
  pages={15522--15533},
  year={2024}
}

@article{li2024imitation,
  title={From Imitation to Exploration: End-to-end Autonomous Driving based on World Model},
  author={Li, Yueyuan and Jiang, Mingyang and Zhang, Songan and Yuan, Wei and Wang, Chunxiang and Yang, Ming},
  journal={arXiv preprint arXiv:2410.02253},
  year={2024}
}

@article{yang2025raw2drive,
  title={Raw2Drive: Reinforcement learning with aligned world models for end-to-end autonomous driving (in carla v2)},
  author={Yang, Zhenjie and Jia, Xiaosong and Li, Qifeng and Yang, Xue and Yao, Maoqing and Yan, Junchi},
  journal={arXiv preprint arXiv:2505.16394},
  year={2025}
}

@article{zheng2024doe,
  title={Doe-1: Closed-loop autonomous driving with large world model},
  author={Zheng, Wenzhao and Xia, Zetian and Huang, Yuanhui and Zuo, Sicheng and Zhou, Jie and Lu, Jiwen},
  journal={arXiv preprint arXiv:2412.09627},
  year={2024}
}

@article{yang2024drivearena,
  title={Drivearena: A closed-loop generative simulation platform for autonomous driving},
  author={Yang, Xuemeng and Wen, Licheng and Ma, Yukai and Mei, Jianbiao and Li, Xin and Wei, Tiantian and Lei, Wenjie and Fu, Daocheng and Cai, Pinlong and Dou, Min and others},
  journal={arXiv preprint arXiv:2408.00415},
  year={2024}
}

@article{shang2025drivedpo,
  title={DriveDPO: Policy Learning via Safety DPO For End-to-End Autonomous Driving},
  author={Shang, Shuyao and Chen, Yuntao and Wang, Yuqi and Li, Yingyan and Zhang, Zhaoxiang},
  journal={arXiv preprint arXiv:2509.17940},
  year={2025}
}

@article{liu2023towards,
    title={Towards socially responsive autonomous vehicles: A reinforcement learning framework with driving priors and coordination awareness},
    author={Liu, Jiaqi and Zhou, Donghao and Hang, Peng and Ni, Ying and Sun, Jian}, journal={IEEE Transactions on Intelligent Vehicles},
    volume={9},
    number={1},
    pages={827--838},
    year={2023},
    publisher={IEEE}
}

\end{document}